\documentclass[a4paper]{cas-sc}

\ExplSyntaxOn
\exp_args:NNno \exp_args:Nno \use:n { \cs_gset:Npn \__make_fig_caption:nn #1#2 }
  {
    \exp_after:wN \use_ii_i:nn \exp_after:wN
      { \__make_fig_caption:nn {#1} {#2} }
      { \dim_set:Nn \l_fig_width_dim \linewidth }
  }
\exp_args:NNno \exp_args:Nno \use:n { \cs_gset:Npn \__make_tbl_caption:nn #1#2 }
  {
    \exp_after:wN \use_ii_i:nn \exp_after:wN
      { \__make_tbl_caption:nn {#1} {#2} }
      { \dim_set:Nn \l_tbl_width_dim \linewidth }
  }
\ExplSyntaxOff

\usepackage[numbers]{natbib}
\usepackage{subfigure}
\usepackage[justification=centering]{caption}
\usepackage{amsmath}
\usepackage{amsthm,amssymb}
\usepackage{arydshln}
\usepackage{verbatim}
\usepackage{graphicx}
\usepackage{subfigure}
\usepackage{booktabs}
\usepackage{amsfonts}
\usepackage{bm}
\usepackage{multirow}
\usepackage{enumerate}

\def\tsc#1{\csdef{#1}{\textsc{\lowercase{#1}}\xspace}}
\tsc{WGM}
\tsc{QE}
\tsc{EP}
\tsc{PMS}
\tsc{BEC}
\tsc{DE}

\begin{document}
\begin{sloppypar}
\let\WriteBookmarks\relax
\def\floatpagepagefraction{1}
\def\textpagefraction{.001}
\shorttitle{MS-LSTM: Exploring Spatiotemporal Multiscale Representations in Video Prediction Domain}
\shortauthors{Ma et~al.}

\title [mode = title]{MS-LSTM: Exploring Spatiotemporal Multiscale Representations in Video Prediction Domain}   



\author[1]{Zhifeng Ma}



\address[1]{Faculty of Computing, Harbin Institute of Technology, Harbin, China}

\author[1]{Hao Zhang}[orcid=0000-0002-6769-2115]
\cormark[1]
\ead{zhh1000@hit.edu.cn}

\author[2]{Jie Liu} 


\address[2]{International Research Institute for Artificial Intelligence, Harbin Institute of Technology, Shenzhen, China}

\cortext[cor1]{Corresponding author}


\begin{abstract}
The drastic variation of motion in spatial and temporal dimensions makes the video prediction task extremely challenging. Existing RNN models obtain higher performance by deepening or widening the model. They obtain the multi-scale features of the video only by stacking layers, which is inefficient and brings unbearable training costs (such as memory, FLOPs, and training time). Different from them, this paper proposes a spatiotemporal multi-scale model called MS-LSTM wholly from a multi-scale perspective. On the basis of stacked layers, MS-LSTM incorporates two additional efficient multi-scale designs to fully capture spatiotemporal context information. Concretely, we employ LSTMs with mirrored pyramid structures to construct spatial multi-scale representations and LSTMs with different convolution kernels to construct temporal multi-scale representations. \textcolor{black}{We theoretically analyze the training cost and performance of MS-LSTM and its components.} Detailed comparison experiments with \textcolor{black}{twelve} baseline models on four video datasets show that MS-LSTM has better performance but lower training costs. 
\end{abstract}


\maketitle

\section{Introduction}
As a fundamental yet essential research task in predictive learning, video prediction or called spatiotemporal predictive learning has aroused widespread research interest in the computer vision community. \textcolor{black}{Video prediction has numerous practical applications in various domains, including robot pose prediction~\cite{xu2018video, wu2021greedy}, precipitation nowcasting~\cite{wei2013soft, yang2022spatio, ma2022focal, espeholt2022deep, ma2023mm}, human trajectory prediction\cite{wang2020probabilistic, chatterjee2021hierarchical, chang2022strpm}, traffic flow prediction~\cite{deng2019exploring, guen2020disentangling, qi2023fedagcn}, and so on.} It is a special case of self-supervision, where the model needs to estimate upcoming future frames from given historical frames at the pixel level. \textcolor{black}{While it is not necessary to have annotated data for training such models, the models need to be able to grasp the intricate dynamics of real-world phenomena (such as physical interactions) in order to create consistent sequences~\cite{castrejon2019improved}.} 

\textcolor{black}{Natural videos are inherently uncertain, and this uncertainty increases over time. To accommodate this uncertainty, the model converges to an average state of the future frame, occurring as visual blurring~\cite{oprea2020review}.} To address this issue, prior approaches choose to deepen or widen the pioneer model. \textcolor{black}{ConvLSTM, which combines convolution with long short-term memory (LSTM) to simultaneously capture spatial and temporal dynamics, has served as a foundational approach upon which subsequent models have been built to improve prediction performance. The evolution of these convolutional recurrent neural networks (ConvRNNs or RNNs) can be broadly traced through ConvLSTM, PredRNN~\cite{wang2017predrnn}, TrajGRU~\cite{shi2017deep}, PredRNN++~\cite{wang2018predrnn++}, MIM~\cite{wang2019memory}, MotionRNN~\cite{wu2021motionrnn}, PredRNN-V2~\cite{wang2022predrnn}, and PrecipLSTM~\cite{ma2022preciplstm}.} These practices of increasing the depth and width of ConvLSTM lead to the increase of model parameters and the expansion of model capacity, which is beneficial to deal with complex nonlinear transformations. Nevertheless, the side effect is the skyrocketing training cost, such as memory (most intractable), FLOPs, and training time, which is unbearable in high-resolution and resource-constrained scenes. 

\textcolor{black}{As \textcolor{black}{pointed} out by Res2Net~\cite{gao2019res2net}, cardinality~\cite{xie2017aggregated} and scale~\cite{gao2019res2net} \textcolor{black}{were} two other essential tricks to improve the performance of neural networks. In view of the fact that grouping convolution can reduce parameters, cardinality uses grouped convolution while increasing the number of groups until the same parameters as before grouping. Regrettably, although cardinality does not increase parameter complexity, it brings extra memory in exchange for performance improvement. Multi-scale representation in vision refers to receptive fields of different sizes, which are used to describe objects of different scales and can be achieved in three ways. (1) Depth. The receptive field of the network will gradually become larger as convolutional layers are stacked, and layers of different depths form a multi-scale representation. For example, VGGNet~\cite{simonyan2014very} and ResNet~\cite{he2016deep}. (2) Kernel. Convolution kernels of different sizes have receptive fields of different sizes, and multiple parallel multi-scale convolution kernels form a multi-scale representation. For example, GoogLeNet~\cite{szegedy2015going}, SKNet~\cite{li2019selective}, and Res2Net~\cite{gao2019res2net}. (3) Downsampling (Pooling). The network usually needs to be stacked deeply to obtain a larger receptive field, but downsampling can double the receptive field~\cite{luo2016understanding}. Generally, pooling layers are inserted between convolutional layers to improve efficiency, which is straightforwardly manifested as a multi-scale representation composed of feature maps of different sizes. For example, LeNet~\cite{lecun1998gradient} and AlexNet~\cite{krizhevsky2017imagenet}. The most effective way among the three is downsampling, where the former two increase the depth and width of the model respectively and increase the overhead, while the latter only increases layers without parameters and can reduce the overhead. The past few years have witnessed a trend of convolutional neural networks (CNNs) towards efficient multi-scale design. Since convolution is an essential part of ConvRNN, how to efficiently construct a multi-scale representation of ConvRNN should also be its future development direction.}

Based on the following two facts: (1) The boom of CNN has witnessed the rapid development of multi-scale technology. (2) A basic problem of video prediction is how to efficiently learn a good spatiotemporal representation of video speculation or inference. In this work, we absorb the essence of the multi-scale design of CNN, combine it with the spatiotemporal characteristics of video, and propose a simple yet efficient spatiotemporal multi-scale approach called multi-scale LSTM (MS-LSTM). Unlike most existing methods that enhance the layer-wise multi-scale representation strength of RNN, MS-LSTM improves the multi-scale representation ability at a more granular level. MS-LSTM incorporates two additional efficient multi-scale designs to fully capture spatiotemporal contextual information. From a spatial perspective, we stack RNN layers like ordinary RNNs and use downsampling to further obtain multi-scale representations. Whereas, video prediction is a pixel reconstruction task. Hence, we add a symmetric decoder to restore the output of the encoder to the original pixel space, resulting in a mirrored pyramid structure like UNet~\cite{ronneberger2015u}. From a temporal perspective, we use LSTMs with different sizes of kernels instead of a single kernel to build multi-scale representations, which we call multi-Kernel LSTM (MK-LSTM). MK-LSTM follows the Markov assumption like ConvLSTM, using different scales of cellular memory to describe the deformation of objects at different scales over time. In addition, we also analyze the training cost (params, memory, FLOPs, time) of MS-LSTM and its components and the reason (increase of receptive field \textcolor{black}{and stepwise generation}) for the performance improvement brought by the multi-scale architecture. It should be noted that the utilization of the big kernel gains rewards but it brings extra training burden, which can partly be resolved by the introduction of downsampling, which can ensure that our model sufficiently learns multi-scale features while bringing modest training cost. Our contributions can be summarized as follows:
\begin{itemize}
\item We propose a new spatiotemporal multiscale model named MS-LSTM, which incorporates three orthogonal multiscale designs.
\item MS-LSTM employs stacked LSTMs with multiple convolution kernels (MK-LSTM) to form a mirrored pyramid structure, which can fully acquire the spatiotemporal context representation and generates video from coarse to fine.
\item We theoretically analyze the training cost and performance of MS-LSTM. Experiments on four datasets with \textcolor{black}{twelve} competing models have proved that MS-LSTM has lower training occupation but higher performance.
\end{itemize}

\textcolor{black}{The remainder of the paper is structured as follows. Section~\ref{related work} briefly reviews related works about various kinds of networks and multi-scale rnn models in the video prediction domain. Section~\ref{preliminaries} formulates the video prediction problem, details the pioneer ConvLSTM, and reviews several baselines (ConvLSTM variants) for comparison. Section~\ref{ms-lstm} presents two crucial modules that constitute MS-LSTM, namely spatial multi-scale LSTM and temporal multi-scale LSTM, and gives detailed training cost and performance analysis about them. Extensive experiments on four datasets in Section~\ref{sec:experiments} demonstrate that MS-LSTM surpasses state-of-the-art CNN, ConvRNN, and Transformer~\cite{vaswani2017attention} models. At last, Section~\ref{conclusion and future work} concludes the paper and anticipates the future.}

\section{Related Work} \label{related work}
\subsection{Video Prediction Models} 
\textcolor{black}{Currently, CNNs, ConvRNNs, probabilistic networks, and Transformer~\cite{vaswani2017attention} networks are four common video prediction models, which introduce different inductive biases when designing the network structure. UNet~\cite{ronneberger2015u} variants are common models for precipitation nowcasting, such as RainNet~\cite{ayzel2020rainnet}, FURENet~\cite{pan2021improving}, \cite{han2021convective}, and Broad-UNet~\cite{fernandez2021broad}. However, UNet has a natural disadvantage in capturing long-term changes and may be more suitable for short-term inference. \textcolor{black}{Recently, SimVP~\cite{gao2022simvp} \textcolor{black}{rehired} UNet for video prediction, which \textcolor{black}{assumed} that convolution (2D by default) can learn spatiotemporal trends simultaneously, but the trick \textcolor{black}{was} training for thousands of epochs.} In contrast, ConvRNN is more famous and has an endless series of variants, for example, ConvLSTM~\cite{shi2015convolutional}, MCNet~\cite{villegas2017decomposing}, TrajGRU~\cite{shi2017deep}, PredRNN~\cite{wang2017predrnn}, PredRNN++~\cite{wang2018predrnn++}, E3D-LSTM~\cite{wang2018eidetic}, MIM~\cite{wang2019memory}, CubicLSTM~\cite{fan2019cubic}, TMU~\cite{yao2020unsupervised}, SA-ConvLSTM~\cite{lin2020self}, PhyDNet~\cite{guen2020disentangling}, \cite{lee2021video}, MotionRNN~\cite{wu2021motionrnn}, MAU~\cite{chang2021mau}, STRPM~\cite{chang2022strpm}, PrecipLSTM~\cite{ma2022preciplstm}, PredRNN-V2~\cite{wang2022predrnn}, and so on. This is because they introduce a reasonable spatiotemporal inductive bias, which makes them easier to train and perfectly fit the task. Generative adversarial networks (GANs)~\cite{mathieu2016deep, wang2020g3an, ravuri2021skilful, huang2022video, li2023future} and variational auto-encoders (VAEs)~\cite{denton2018stochastic, wang2020probabilistic, akan2021slamp, wu2021greedy} \textcolor{black}{learned} the underlying distribution of video data by optimizing the KL divergence in loss functions, which is why we call them probabilistic networks. Transformer~\cite{vaswani2017attention} \textcolor{black}{was} born for massive time series data. It will be data hungry and hard to train without introducing spatiotemporal inductive bias. \textcolor{black}{Compared to VideoGPT~\cite{yan2021videogpt}, TATS~\cite{ge2022long}, and MaskViT~\cite{gupta2022maskvit}, Earthformer~\cite{gao2022earthformer} and MIMO-VP~\cite{ning2023mimo} \textcolor{black}{were} more suitable for video prediction tasks, because they \textcolor{black}{introduced} local spatiotemporal (cuboid attention and 3D convolution) prior assumptions.} This paper proposes a new network belonging to ConvRNN, and we compare it with other advanced CNN, ConvRNN, and Transformer networks.}

\subsection{Multiscale ConvRNN Models} 
\textcolor{black}{Two multiscale ConvRNNs were recently published, both of which are potential competitors. Both CMS-LSTM~\cite{chai2022cms} and \textcolor{black}{MoDeRNN~\cite{chai2022modernn}} contributed two innovations, context embedding and multi-scale expression. Both contextual embeddings used the form of mutual gating of current input and previous output, but their multi-scale designs were quite different, where CMS-LSTM was multi-scale attention \textcolor{black}{while MoDeRNN was multi-scale kernels.} However, attention leads to quadratic complexity, \textcolor{black}{and large convolution kernels are expensive.} Moreover, experiments in Section~\ref{sec:taxi-ms-vs} show that their multi-scale designs are not only expensive but also inefficient.}

\section{Preliminaries}\label{preliminaries}
\subsection{Problem Formulation} \label{sec:problem formulation}
\textcolor{black}{Video prediction is a spatiotemporal prediction task. In terms of space, it is a frame at a certain moment. We use $X_t \in \mathbb{R}^{c \times h \times w}$ to represent it, where $t$ is time while $c$, $h$, and $w$ represent the channel, height, and width of the frame respectively. In terms of time, it is a time sequence \{$X_0, ..., X_{m-1}, X_{m}, ..., X_{m+n-1}$\} composed of multiple frames, where $m+n$ is the sequence length. The video prediction task is to predict unknown $n$ frames $Y=\{X_{m}, ..., X_{m+n-1}$\} given known $m$ frames $X=\{X_0, ..., X_{m-1}$\}. This problem is generally solved using maximum likelihood estimation:
\begin{gather}\label{eq::problem}
    \begin{split}
        & \theta^* = \mathop{\arg\max}\limits_{\theta} \; P(Y|X;\theta). \\
	\end{split}
\end{gather} 
where $\theta$ is the parameters of the neural network, and $\theta^*$ is the optimal parameters, which are optimized by the stochastic gradient descent algorithm.}

\begin{figure}[htbp]
	\centering
	\includegraphics[width=0.32\linewidth]{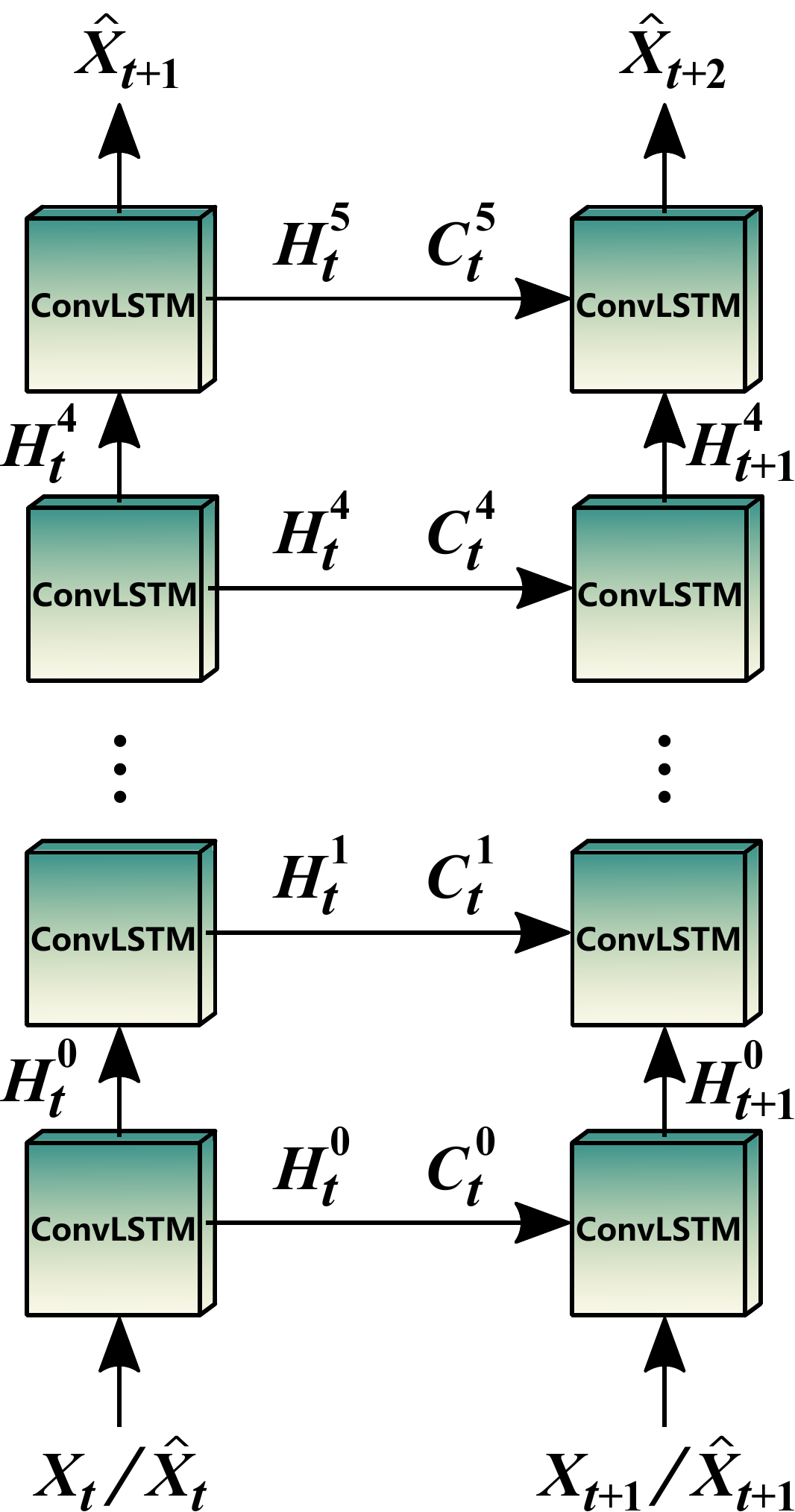}
	\caption{\textcolor{black}{The architecture of ConvLSTM. It only uses depth to obtain the multi-scale representation.}}
	\label{fig::convlstm}
\end{figure}

\subsection{ConvLSTM} \label{sec:convlstm} 
\textcolor{black}{In 2015, Shi et al. \textcolor{black}{combined} the convolution suitable for spatial tasks and the LSTM~\cite{hochreiter1997long} suitable for temporal tasks to propose ConvLSTM~\cite{shi2015convolutional}. ConvLSTM can simultaneously capture spatiotemporal motion and perfectly adapts to video prediction tasks. This characteristic also establishes its status as the father of ConvRNN. The formula of the ConvLSTM unit~\cite{shi2015convolutional, lin2020self} is
\begin{equation}
    \begin{aligned}
    & i_t = \sigma(W_{ix} \ast X_{t}^l + W_{ih} \ast H_{t-1}^l), \\
    & f_t = \sigma(W_{fx} \ast X_{t}^l + W_{fh} \ast H_{t-1}^l), \\
    & g_t = \tanh(W_{gx} \ast X_{t}^l + W_{gh} \ast H_{t-1}^l), \\
    & C_{t}^l = f_t \odot C_{t-1}^l + i_t \odot g_t, \\
    & o_t = \sigma(W_{ox} \ast X_{t}^l + W_{oh} \ast H_{t-1}^l), \\
    & H_{t}^l = o_t \odot \tanh(C_{t}^l).
    \label{eq::convlstm}
    \end{aligned}
\end{equation}
$t$ and $l$ represent time and layer respectively. $i$, $f$, $g$, and $o$ denote input, forget, modulation, and output gates respectively. $X$ is the input, $H$ is the hidden state, $C$ is the cell state, and $W$ is the parameter. The sigmoid and tanh activation functions are expressed in terms of $\sigma$ and $\tanh$ respectively. Convolutions and Hadamard products are marked with $\ast$ and $\odot$ respectively. Fig.~\ref{fig::convlstm} shows the ConvLSTM architecture stacked with 6-layer units.}

\begin{table}[htbp]
    \centering
	\caption{\textcolor{black}{The complexity of the convolution and ConvLSTM.}}
	\resizebox{0.42\linewidth}{!}{
		\centering
		\begin{tabular}{lccccc}
			\toprule
			\centering
            Models & Params & FLOPs & Memory \\
            \midrule
            Convolution & $c^2k^2$ & $2bc^2hwk^2$ & $bchw$ \\
            ConvLSTM & $8c^2k^2$ & $16bc^2hwk^2$ & $16bchw$ \\
            \bottomrule
		\end{tabular}
	} 
	\label{table:convlstm-complexity}
    \vspace{-0.4cm}
\end{table}

\textcolor{black}{Next, we will analyze the complexity of ConvLSTM based on convolution to prepare for the subsequent sections. We first summarize the parameters, computation (FLOPs), and space (memory) complexity  of the two in Table~\ref{table:convlstm-complexity}, where $k$, $b$, $c$, $h$, and $w$ represent the convolution kernel size, batch size, channel, height, and width, respectively. Here are some assumptions: Assume that only one layer of convolution and one ConvLSTM unit are studied. Assume the input and output channels are the same ($c$). Assume that the input and output features are of the same size ($h \times w$). Assume that the size of the output tensor is used to represent the space complexity. Specifically, the parameters, FLOPs, and memory of convolution are easy to obtain, which are $c^2k^2$, $2bc^2hwk^2$, and $bchw$, respectively. Since the ConvLSTM unit is equipped with 8 convolutions, its parameters and FLOPs are $8c^2k^2$ and $16bc^2hwk^2$ respectively, where we ignore relatively small FLOPs other than convolutions~\cite{pfeuffer2019separable}. $+$ does not take up any memory~\cite{sohoni2019low}, but activation layers and Hadamard layers have the same memory footprint as convolutions. The number of these memory-consuming operations is 16 in total, so the memory of ConvLSTM is $16bchw$.}

\subsection{ConvLSTM Variants} \label{sec:convlstm variants} 

\textcolor{black}{In order to enhance the video prediction ability of ConvLSTM~\cite{shi2015convolutional}, previous ConvRNNs \textcolor{black}{embarked} on the road of widening (TrajGRU, PredRNN, MIM, PredRNN-V2, PrecipLSTM) and deepening (PredRNN++, MotionRNN) the base model. They \textcolor{black}{employed} a recursive prediction architecture similar to the ConvLSTM in Fig.~\ref{fig::convlstm}. Shi et al. \textcolor{black}{integrated} the traditional optical flow method into ConvGRU~\cite{shi2017deep} to propose TrajGRU~\cite{shi2017deep}, in which recursive connections \textcolor{black}{were} dynamically determined over time. On the basis of ConvLSTM, Wang et al. \textcolor{black}{introduced} a spatiotemporal memory along the zigzag propagation to construct PredRNN~\cite{wang2017predrnn}. Wang et al. \textcolor{black}{redefined} the formula of PredRNN and \textcolor{black}{introduced} a gradient highway unit (GHU), which \textcolor{black}{was} called PredRNN++~\cite{wang2018predrnn++}. Wang et al. \textcolor{black}{believed} that limited differences can bring a stationary process. They \textcolor{black}{added} stationary memory and non-stationary memory to launch MIM~\cite{wang2019memory} on the basis of PredRNN. Wu et al. \textcolor{black}{borrowed} from the momentum update principle of the optimizer and \textcolor{black}{designed} MotionGRU~\cite{wu2021motionrnn} that can memorize short-term and long-term motion trends. They \textcolor{black}{built} MotionRNN~\cite{wu2021motionrnn} by embedding MotionGRU between the~\cite{wu2021motionrnn} layers of MIM. Ma et al. \textcolor{black}{introduced} PrecipLSTM~\cite{ma2022preciplstm} for precipitation nowcasting based on PredRNN, which \textcolor{black}{invented} meteorological spatiotemporal memories based on the first law of geography. PredRNN-V2~\cite{wang2022predrnn} \textcolor{black}{disentangled} the two memories in PredRNN and \textcolor{black}{designed} a reverse scheduled sampling strategy, which \textcolor{black}{was} also the latest work of Wang et al. Although these methods of deepening and widening the model bring about performance improvements, this is at the expense of a huge training cost, which makes them difficult to apply to high-resolution tasks. Models with high performance and low resource consumption may be more popular. }

\begin{figure}[htbp]
	\centering
	\includegraphics[width=0.6\linewidth]{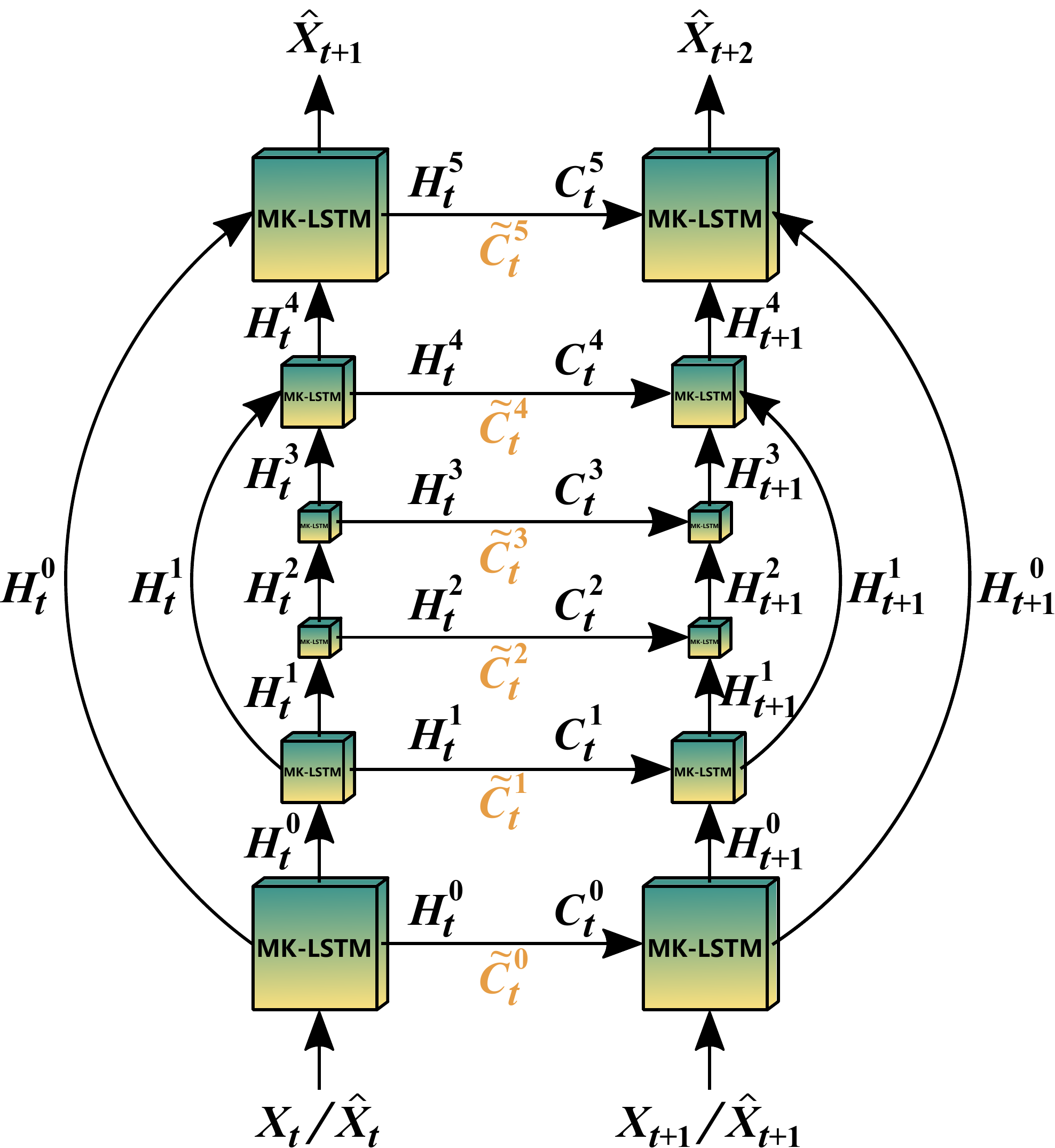}
	\caption{The architecture of MS-LSTM. It uses depth, downsampling, and multiple kernels to obtain multi-scale representations. The model performs one-step predictions along the spatial axis (vertical direction) while passing hidden states between layers. The model extrapolates future frames along the time axis (horizontal direction) while passing the hidden and cell states over time. Skip connections (``+'') are represented by curves to combine the features of the encoder and decoder at the same scale, enabling the model to generate static or high-frequency features easily~\cite{denton2018stochastic}. The yellow symbols $\tilde{C}_{t}$ represent the newly added multi-scale cell memory. }
	\label{fig::ms-lstm}
\end{figure}

\section{Multi-Scale LSTM} \label{ms-lstm}

Due to the high dimensionality and uncertainty of natural videos, extracting a robust representation from raw pixel values is an overly complicated task. The per-pixel variability between consecutive frames causes exponential growth in the prediction error on the long-term horizon~\cite{oprea2020review}. Common models struggle with blurriness, and pixel-level video prediction is still challenging. 

A straightforward idea is that if we hope to foretell the future, we need to memorize as many historical events as possible. When we recall something that happened before, we do not just recall object movements, but also recollect visual appearances from coarse to fine~\cite{wang2017predrnn}. Motivated by this, we present a new recurrent architecture called multi-scale LSTM (MS-LSTM), which combines the multi-scale design of traditional CNNs with the spatiotemporal motion property of videos. MS-LSTM (6 layers, Fig.~\ref{fig::ms-lstm}) integrates three orthogonal multi-scale designs, which are depth, downsampling, and multiple convolution kernels, where the first two constitute spatial multi-scale LSTM (SMS-LSTM), and the first and last constitute temporal multi-scale LSTM (TMS-LSTM). The use of multi-scale features makes it easy to capture the motion of objects at different scales, where the small scale focuses on the contour and shape transformation of the object, while the large scale focuses on detailed features, such as the trajectories of human limbs or the evolution of precipitation. We experimentally verify that MS-LSTM has strong spatiotemporal modeling capabilities but consumes fewer training resources (Section~\ref{sec:experiments}). We'll cover the specifics below.



\subsection{Spatial Multi-Scale LSTM} \label{sms-lstm}

Limited by the scale of the convolution kernel, convolution can only capture the short-distance spatial dependence in the image~\cite{mathieu2016deep}. One way is to use stacked convolution layers. The other way is to use the pooling operation. Both have brought about an increase in the spatial receptive field, making the model see wider, which laid the foundation for today's CNN. The ConvRNNs (like PredRNN++~\cite{wang2018predrnn++} and MotionRNN~\cite{wu2021motionrnn}) in the video prediction domain have absorbed the essence of ``deep” learning, and most of them adopt the method of stacking more modules to obtain a wider spatiotemporal receptive field. However, unlike convolution operations, recursive operations cannot be parallelized, resulting in a surge in memory usage and training time. Therefore, blindly increasing the depth will no longer be the most effective way. The use of pooling will allow the model to better sense contextual information and reduce training requirements by working at low resolutions. Regrettably, the use of downsampling to capture spatial dependencies has been poorly explored among ConvRNNs. 

In this paper, we improve the existing RNN models by using the familiar approaches in CNN, stacking ConvRNN layers but interspersing downsampling. Yet using downsampling will bring about a loss of image resolution, which is unacceptable for the pixel-level prediction task that requires the output to have the same resolution as the input. Inspired by UNet~\cite{ronneberger2015u}, we introduce a symmetric decoder to restore the output of the encoder to the original pixel space, resulting in a mirrored pyramid structure. Specifically, supposing we stack 6 layers of ConvRNN (Fig.~\ref{fig::ms-lstm}), then the bottom three layers make up the encoder and the top three layers make up the decoder. We gradually decrease the resolution of the encoder but gradually increase the resolution of the decoder. The downsampling and upsampling operations are implemented by max pooling and bilinear interpolation, respectively. Besides, we add skip connections (``+'') between the encoder and decoder at the same scale to enable the model to generate static \textcolor{black}{or high-frequency} features easily~\cite{denton2018stochastic}. Ultimately, we obtain a spatial multi-scale LSTM. 

It should be emphasized that the LSTM unit can be replaced by the unit of existing models, such as ConvLSTM, TrajGRU, PredRNN, PredRNN++, MIM, MotionRNN, PrecipLSTM, etc. Given that their basic units are either too complex (purely for novelty) or do not introduce multi-scale design, we redesign a new ConvRNN unit (MK-LSTM) that is relatively simple and introduces the temporal multi-scale design, which will be presented in the next section. The SMS-LSTM in subsequent parts of this paper refers to the one that uses the ConvLSTM unit. 

\begin{figure}[htbp]
	\centering
	\includegraphics[width=0.6\linewidth]{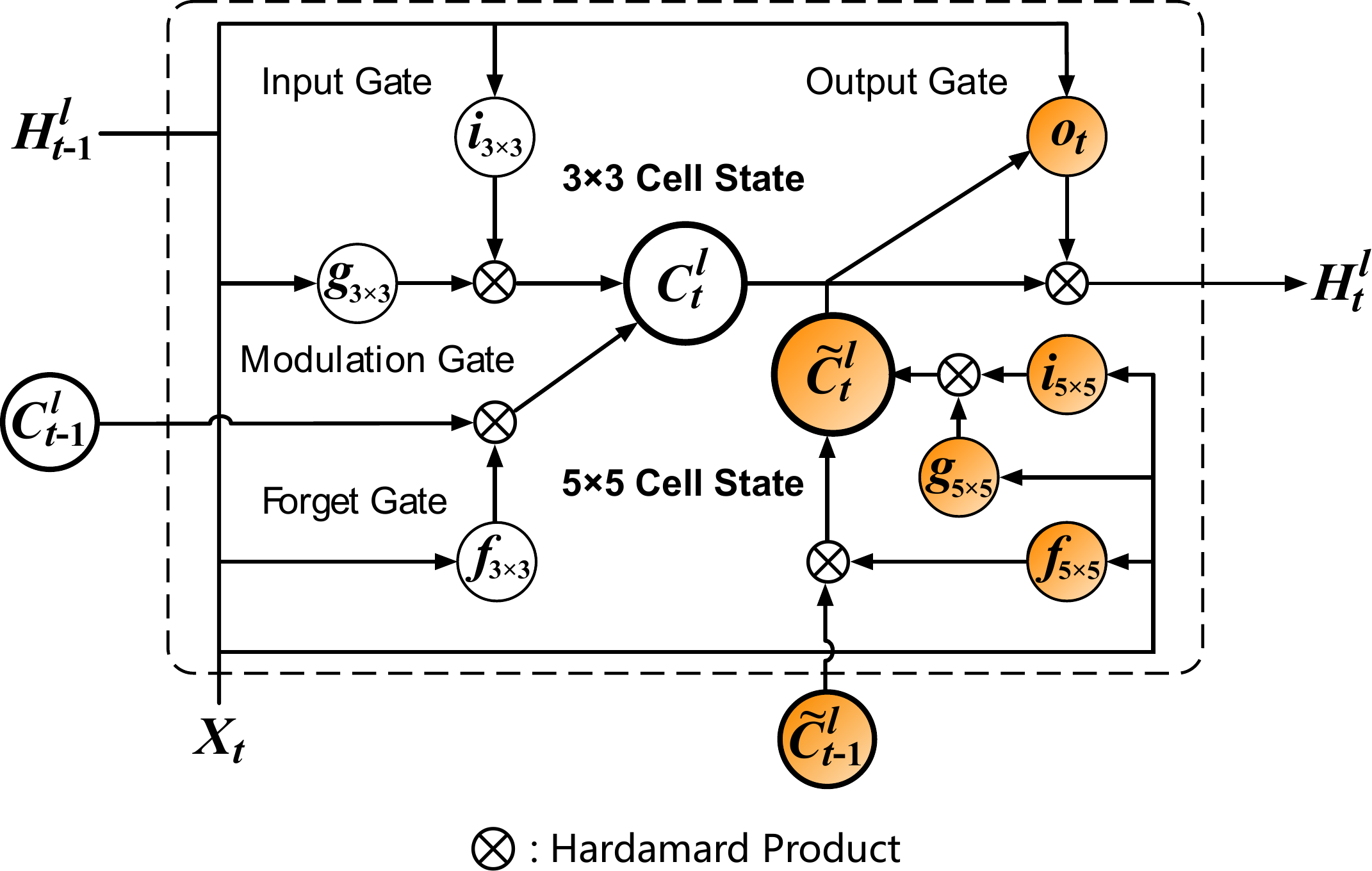}
	\caption{\textcolor{black}{The architecture of MK-LSTM. Orange-filled circles denote the differences between MK-LSTM and ConvLSTM (White-filled circles).}}
	\label{fig::mk-lstm}
\end{figure}

\subsection{Temporal Multi-Scale LSTM} \label{tms-lstm}

In addition to stacking layers and using pooling operations, GoogLeNet~\cite{szegedy2015going} also \textcolor{black}{utilized} the Inception composed of parallel filters of different sizes to enhance the multiscale representation capability. It also \textcolor{black}{motivated} some follow-up work. In theory, superimposing two convolutional layers with $3\times3$ kernels is equivalent to a convolutional layer with $5\times5$ kernel, Res2Net~\cite{gao2019res2net} \textcolor{black}{formed} an equivalent multi-kernel layer like GoogLeNet by adding shortcuts between parallel group convolutions. SKNet~\cite{li2019selective} \textcolor{black}{proposed} a dynamic selection mechanism that can select $3\times3$ or $5\times5$ kernel in a soft-attention manner. Inspired by these works, we design a multi-scale convolutional LSTM to explore fine-grained temporal multiscale representations, which we call multi-kernel LSTM (MK-LSTM). The temporal multi-scale LSTM (TMS-LSTM) will be obtained by stacking multiple MK-LSTM units \textcolor{black}{(6 layers)}. The intuitive view of TMS-LSTM is to replace ConvLSTM units in Fig.~\ref{fig::convlstm} with MK-LSTM units.

MK-LSTM (Fig.~\ref{fig::mk-lstm} and Eq.~(\ref{eq::mk-lstm})) follows the Markov assumption like ConvLSTM, using different scales of cellular memory to describe the deformation of objects at different scales over time. We only employ two types of convolution kernels: $3\times3$ and $5\times5$. There are two reasons: (1) For the $1\times1$ kernel, the receptive field will not grow as layers stack. (2) \textcolor{black}{For the kernel larger than $1\times1$, the deeper layer has larger receptive fields.} However, using an excessively large convolution kernel will lead to a rapid increase in training consumption. Based on the above reasons, we construct MK-LSTM using two ConvLSTMs with $3\times3$ and $5\times5$ kernel respectively, a compromise between performance and consumption. The equation of the MK-LSTM unit is shown as follows:
\begin{equation}
    \begin{aligned}
    & i_{3 \times 3} = \sigma(W_{ix} \ast X_{t}^l + W_{ih} \ast H_{t-1}^l), \\
    & f_{3 \times 3} = \sigma(W_{fx} \ast X_{t}^l + W_{fh} \ast H_{t-1}^l), \\
    & g_{3 \times 3} = \tanh(W_{gx} \ast X_{t}^l + W_{gh} \ast H_{t-1}^l), \\
    & C_{t}^l = f_{3 \times 3} \odot C_{t-1}^l + i_{3 \times 3} \odot g_{3 \times 3}, \\ 
    & i_{5 \times 5} = \sigma(W_{ix}^{'} \ast X_{t}^l + W_{ih}^{'} \ast H_{t-1}^l), \\
    & f_{5 \times 5} = \sigma(W_{fx}^{'} \ast X_{t}^l + W_{fh}^{'} \ast H_{t-1}^l), \\
    & g_{5 \times 5} = \tanh(W_{gx}^{'} \ast X_{t}^l + W_{gh}^{'} \ast H_{t-1}^l), \\
    & \tilde{C}_{t}^l = f_{5 \times 5} \odot \tilde{C}_{t-1}^l + i_{5 \times 5} \odot g_{5 \times 5}, \\
    & o_t = \sigma(W_{ox} \ast X_{t}^l + W_{oh} \ast H_{t-1}^l + W_{oc} \ast C_{t}^l \\
    & \hspace{6.3mm} + W_{ox}^{'} \ast X_{t}^l + W_{oh}^{'} \ast H_{t-1}^l + W_{oc}^{'} \ast \tilde{C}_{t}^l), \\
    & H_{t}^l = o_t \odot \tanh(W_{1 \times 1} \ast [C_{t}^l, \tilde{C}_{t}^l]), 
    \label{eq::mk-lstm}
    \end{aligned}
\end{equation}
\textcolor{black}{where $W$ and $W^{'}$ are the parameters to be optimized; $C$ and $\tilde{C}$ represent cell states of different kernel size; $i_{*\times*}$, $f_{*\times*}$, and $g_{*\times*}$ stand for input gate, forget gate, and modulation gate of different kernel size respectively; $o$ stands for output gate; \textcolor{black}{$[]$ indicates the channel connection.}}

MK-LSTM adds additional multi-scale cellular memory on top of ConvLSTM, which makes it simple for the model to recall past object changes at different scales and perform long-term predictions. Furthermore, using different convolution kernels results in different sizes of receptive fields, which makes it easy to capture motion at different speeds, where the larger convolution kernel can capture faster motion while the smaller convolution kernel can capture slower motion~\cite{shi2015convolutional}. This characteristic has been confirmed in experiments (Section~\ref{sec:experiments}). From the production of the Moving MNIST~\cite{srivastava2015unsupervised} and KTH~\cite{schuldt2004recognizing} datasets, it can be intuitively seen that there are different rates of movement. The TaxiBJ~\cite{zhang2017deep} and Germany~\cite{ayzel2020rainnet} datasets implicitly contain different rates of movement, \textcolor{black}{because the movement speed of traffic and precipitation should also be random.} It should be noted that the utilization of multiple kernels \textcolor{black}{(width)}, likes depth, gains rewards but brings extra training burden, which can partly be resolved by the introduction of downsampling in SMS-LSTM, which ensures that MS-LSTM can thoroughly learn multi-scale representations without causing excessive training cost. 


\begin{table}[htbp]
    \centering
	\caption{\textcolor{black}{The complexity of the ConvLSTM and MK-LSTM units.}}
	\resizebox{0.4\linewidth}{!}{
		\centering
		\begin{tabular}{lcccc}
            \toprule
            \centering
            Models & Params & FLOPs & Memory \\
            \midrule
            ConvLSTM & $72c^2$ & $144bc^2hw$ & $16bchw$ \\
            MK-LSTM & $307c^2$ & $614bc^2hw$ & $32bchw$ \\
            \bottomrule
		\end{tabular}
	} 
	\label{table:unit-complexity}
\end{table}

\subsection{Analysis of Training Cost and Performance} \label{interprete}
\subsubsection{Analysis of Training Cost} \label{analysis cost}

\textcolor{black}{From Section~\ref{sec:convlstm}, we know that the parameter complexity of a ConvLSTM unit (kernels $k=3$) is $72c^2$, the computation complexity (FLOPs) of a ConvLSTM unit is $144bc^2hw$, and the space complexity (memory) of a ConvLSTM unit is $16bchw$. Analogously, we can extend these training cost representations of ConvLSTM units to MK-LSTM units. The parameter complexity of a MK-LSTM unit (kernels $k=1, 3, 5$) is $307c^2$, the computation complexity (FLOPs) of a MK-LSTM unit is $614bc^2hw$, and the space complexity (memory) of a MK-LSTM unit is $32bchw$. We display them in Table~\ref{table:unit-complexity}.}
 
\textbf{Params Analysis}. Since the parameters of the ConvLSTM (MK-LSTM) unit are independent of the scale ($hw$) and the max pooling layer, the bilinear interpolation layer, and skip connections (``+'') will not increase the parameters of the model, SMS-LSTM (MS-LSTM) has the same parameters as ConvLSTM (TMS-LSTM). In addition, TMS-LSTM (MS-LSTM) has more parameters than ConvLSTM (SMS-LSTM), \textcolor{black}{which is because the MK-LSTM unit has more parameters than the ConvLSTM unit.}

\textbf{Memory Analysis}. During the model training process, there are mainly three parts of memory usage, model memory (\textcolor{black}{$M_\mathrm{par}$}), optimizer memory (3\textcolor{black}{$M_\mathrm{par}$} for Adam~\cite{kingma2014adam}), and output memory (2\textcolor{black}{$M_\mathrm{out}$})~\cite{sohoni2019low}. The model memory is used to store model parameters, the optimizer memory is used to store the gradient and momentum buffer of the parameter, and the output memory consists of two equal parts, which are the forward output memory (\textcolor{black}{$M_\mathrm{par}$}) to store the layer output and the backward output memory to store the output gradient~\cite{gao2020estimating}. In summary, the total memory footprint of the model during training \textcolor{black}{$M_\mathrm{all}=4M_\mathrm{par} + 2M_\mathrm{out}$}. 

Since SMS-LSTM (MS-LSTM) has the same parameters as ConvLSTM (TMS-LSTM), SMS-LSTM (MS-LSTM) has the same model memory (\textcolor{black}{$M_\mathrm{par}$}) as ConvLSTM (TMS-LSTM). However, the output memory of a ConvLSTM (MK-LSTM) unit is proportional to the scale ($hw$). SMS-LSTM (MS-LSTM) uses small-scale ConvLSTM (MK-LSTM) units, so it has fewer forward output memory (\textcolor{black}{$M_\mathrm{out}$}) than ConvLSTM (TMS-LSTM), that is, fewer total memory usage \textcolor{black}{$M_\mathrm{all}$}. In addition, TMS-LSTM (MS-LSTM) has more total memory footprint than ConvLSTM (SMS-LSTM), \textcolor{black}{which is because the MK-LSTM unit has more parameters ($M_\mathrm{par}$) and more layer output memory ($M_\mathrm{out}$) than the ConvLSTM unit.} 

\textbf{FLOPs Analysis}. Since the FLOPs of a ConvLSTM (MK-LSTM) unit is proportional to the scale ($hw$) and SMS-LSTM (MS-LSTM) uses small-scale ConvLSTM (MK-LSTM) units, \textcolor{black}{supposing neglecting the FLOPs of few sampling and skip connection (``+'') operations}, SMS-LSTM (MS-LSTM) has fewer FLOPs than ConvLSTM (TMS-LSTM). In addition, TMS-LSTM (MS-LSTM) has more FLOPs than ConvLSTM (SMS-LSTM), \textcolor{black}{which is because the MK-LSTM unit has more FLOPs than the ConvLSTM unit.}

\textbf{Time Analysis}. 
Theoretically, the smaller the feature size, the fewer times to calculate the convolution, and the lower the time complexity. However, in practical applications, convolution operations are generally converted to matrix operations (img2col in Caffe~\cite{jia2014caffe} or unfold in Pytorch~\cite{paszke2019pytorch}) to speed up calculations, and the current optimization technology for matrix operations is very mature, which eventually leads to the time complexity of ConvLSTM \textcolor{black}{(MK-LSTM)} being insensitive to scale ($hw$). Therefore, the time complexity of SMS-LSTM (MS-LSTM) is not much different from ConvLSTM (TMS-LSTM). But the time complexity of TMS-LSTM (MS-LSTM) is larger than ConvLSTM (SMS-LSTM) due to the fact that \textcolor{black}{the MK-LSTM unit has more convolution layers with larger kernels than the ConvLSTM unit.} 

\begin{figure}[htbp]
	\centering
	\includegraphics[width=0.7\linewidth]{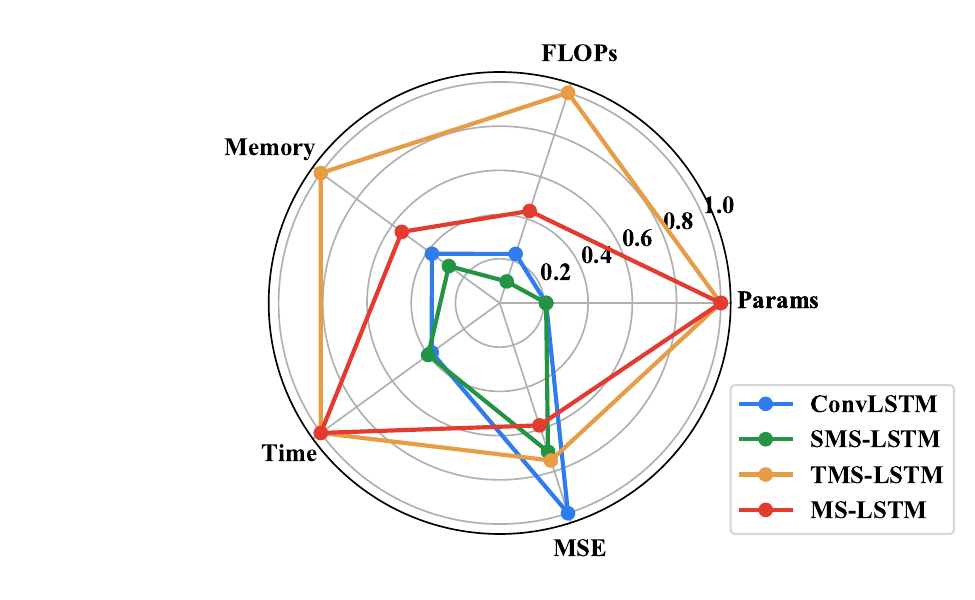}
	\caption{\textcolor{black}{The training cost (params, FLOPs, memory, and time) and performance (MSE) comparison between ConvLSTM, SMS-LSTM, TMS-LSTM, and MS-LSTM. It is a normalized version of the data in Table~\ref{table:mnist-ablation}. For these five indicators, the closer the model is to the center, the better.}}
	\label{fig:radar}
\end{figure}

\textcolor{black}{In summary,} compared with ConvLSTM (TMS-LSTM), SMS-LSTM (MS-LSTM) has the same parameters and training time but less memory and FLOPs. Compared with ConvLSTM (SMS-LSTM), TMS-LSTM (MS-LSTM) has larger parameters, training time, memory, and FLOPs; compared with ConvLSTM, MS-LSTM has larger parameters, moderate training time, moderate memory, and moderate FLOPs because of the neutralizing effect of SMS-LSTM. These conclusions are verified in Section~\ref{sec:ablation}. \textcolor{black}{In addition, we also visualize all the above conclusions with a radar chart (Fig~\ref{fig:radar}), which also indicates that MS-LSTM has moderate training cost but the best performance.}

\subsubsection{Analysis of Performance} \label{analysis performance}
\textcolor{black}{To the best of our knowledge, there are few works on the interpretability of ConvRNN for video prediction, among which \cite{huang2022understanding} is relatively systematic and comprehensive. \cite{huang2022understanding} tried to understand the video prediction task from the perspective of mathematical theory and visualization experiments, which held that video prediction was a coarse-to-fine generation procedure (decoder),} depending on the extending the present and erasing the past theory to catch spatiotemporal motion (encoder). \textcolor{black}{Fortunately, there \textcolor{black}{were} lots of works~\cite{luo2016understanding, ding2022scaling, gu2021interpreting} on the interpretability of CNN in other domains, which \textcolor{black}{were} also based on codec theory, where convolutions and poolings (optional) \textcolor{black}{constituted} the encoder and full connections \textcolor{black}{constituted} the decoder. They \textcolor{black}{tried} to analyze and print the receptive field of the encoder to explain how models work. Since ConvRNN is also mainly composed of convolutions, we also learn from these experiences. In short, we mainly study the encoder of ConvRNN from the perspective of the receptive field, and we believe that the decoder should be a natural decoding process from coarse to fine, as analyzed in \cite{huang2022understanding}.}


\textcolor{black}{Assumptions before analysis: The models all use 6-layer ConvRNN, where the first 3 layers form the encoder and the last 3 layers form the decoder. ConvLSTM and SMS-LSTM use $3 \times 3$ convolution kernels, while TMS-LSTM and MS-LSTM use $3 \times 3$ and $5 \times 5$ convolution kernels. Activation functions ($\sigma$ and $\tanh$) and element-wise operations ($+$ and $\odot$) do not change the receptive field. A ConvLSTM unit should have the same receptive field as a convolution layer in it (Eq.~(\ref{eq::convlstm})), such as $W_{*x} \ast X_{t}^l$ or $W_{*h} \ast H_{t-1}^l$. Then the theoretical receptive field of the ConvLSTM encoder is $7 \times 7$. Since the pooling layer can double the receptive field~\cite{luo2016understanding}, the theoretical receptive field of the SMS-LSTM encoder is $15 \times 15$.} \textcolor{black}{Analogously, the theoretical receptive field of the MS-LSTM encoder will be approximately twice that of TMS-LSTM.} On the basis of SMS-LSTM, we replace the ConvLSTM unit in it with the MK-LSTM unit using a large convolution kernel to continually increase the receptive field. This leads to the receptive field of the MS-LSTM encoder being much larger than $15 \times 15$. \textcolor{black}{There is no doubt that the model with a wider spatiotemporal receptive field will have a stronger ability to simulate complex spatiotemporal changes. The multi-scale decoder of SMS-LSTM (MS-LSTM) decodes progressively from low-resolution to high-resolution, while the decoder of ConvLSTM (TMS-LSTM) does not have this natural coarse-to-fine process. In addition, multi-scale decoders enjoy the benefits of skip connections, which bring multi-scale features of encoders to facilitate decoding high-frequency information.}

\textcolor{black}{The above analysis of the receptive field of the encoder and the multi-scale decoder is presented from a spatial perspective. However, the video prediction task is a spatiotemporal prediction problem, which can also be analyzed from a temporal perspective. Compared with ConvLSTM (SMS-LSTM) using only a single memory cell, TMS-LSTM (MS-LSTM) using more memory cells can remember longer spatiotemporal motion, which is why previous models (like PredRNN~\cite{wang2017predrnn}, MIM~\cite{wang2019memory}, and PrecipLSTM~\cite{ma2022preciplstm}) choose to use more memory cells to widen the model for higher performance.}

\textcolor{black}{In summary, the encoder of SMS-LSTM (MS-LSTM) has a larger spatiotemporal receptive field than the encoder of ConvLSTM (TMS-LSTM), and the decoder of SMS-LSTM (MS-LSTM) enjoys the benefit of skip connections and decodes from coarse to fine, which leads to SMS-LSTM (MS-LSTM) outperforming ConvLSTM (TMS-LSTM). In addition, the performance of TMS-LSTM (MS-LSTM) is better than ConvLSTM (SMS-LSTM), because the MK-LSTM unit has a wider receptive field (larger convolution kernel) and an additional memory cell ($\tilde{C}_t^l$) than the ConvLSTM unit.}

\textcolor{black}{These conclusions are verified in Section~\ref{sec:visual}. In detail, we print the output of each layer of the MS-LSTM, SMS-LSTM, TMS-LSTM, and ConvLSTM. In addition, we also visualize all the above conclusions with a radar chart (Fig~\ref{fig:radar}), which also indicates that MS-LSTM has moderate training cost but the best performance.}

\begin{figure*}[htbp]
	\centering
	\includegraphics[width=1\linewidth]{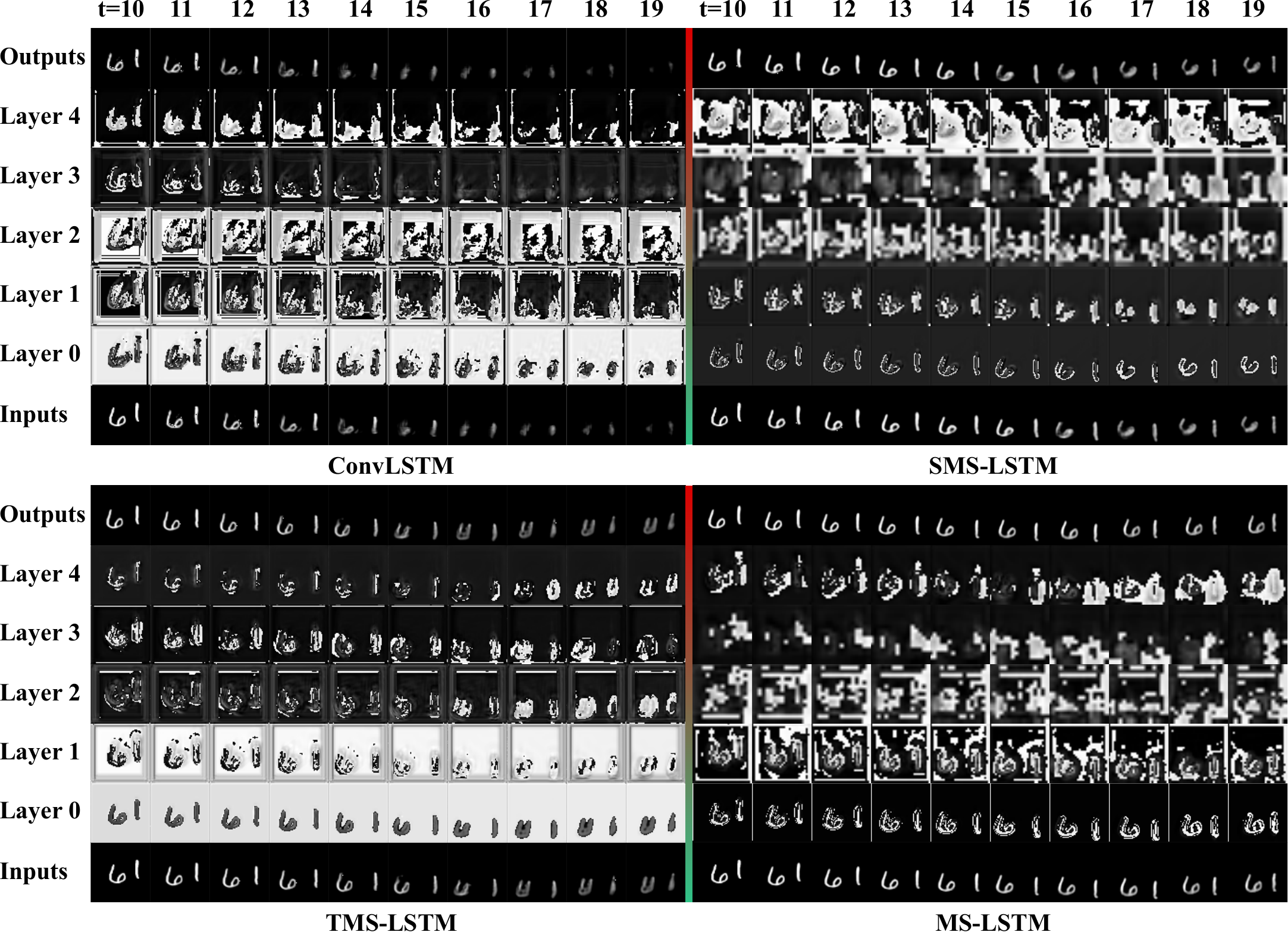}
	\caption{\textcolor{black}{The layer outputs of ConvLSTM, SMS-LSTM, TMS-LSTM, and MS-LSTM on the Moving MNIST dataset.}}
	\label{fig:minist_layer_out_0}
	\vspace{-0cm}
\end{figure*}

\section{Experiments}\label{sec:experiments} 
\textcolor{black}{We compare our model MS-LSTM with 12 baseline models on 4 datasets. The datasets include Moving MNIST~\cite{srivastava2015unsupervised}, TaxiBJ~\cite{zhang2017deep}, KTH~\cite{schuldt2004recognizing}, and Germany~\cite{ayzel2020rainnet}. The baseline models include ConvLSTM~\cite{shi2015convolutional}, TrajGRU~\cite{shi2017deep}, PredRNN~\cite{wang2017predrnn}, PredRNN++~\cite{wang2018predrnn++}, MIM~\cite{wang2019memory}, MotionRNN~\cite{wu2021motionrnn}, PrecipLSTM~\cite{ma2022preciplstm}, PredRNN-V2~\cite{wang2022predrnn}, CMS-LSTM~\cite{chai2022cms}, MoDeRNN~\cite{chai2022modernn}, SimVP~\cite{gao2022simvp}, and Earthformer~\cite{gao2022earthformer}. The results all prove the superiority of MS-LSTM in training cost and performance.}

\subsection{Implementation Details} \label{sec:imple}
\textcolor{black}{The experimental equipment in this article is a machine equipped with 4 NVIDIA A100 40G graphics cards. We use distributed data parallel technology~\cite{paszke2019pytorch} to alleviate the tension of video memory. Specifically, we use a batch size of 4, and each card is only responsible for training one batch. They automatically synchronize the gradient to complete the entire training. Since each card is load balanced, we only need to record the memory of any card. All ConvRNNs have the same experimental configuration, and non-RNNs refer to the official settings (Section~\ref{non-rnn}). The layers, optimizer, initial learning rate, loss, and channels of ConvRNN are 6, Adam~\cite{kingma2014adam}, 0.0003, $L_1 + L_2$, 32, respectively. All ConvRNNs use $3 \times 3$ convolution kernels, except MS-LSTM ($3 \times 3$ and $5 \times 5$) and MoDeRNN ($3 \times 3$, $5 \times 5$, and $7 \times 7$). Moving MNIST, TaxiBJ, KTH, and Germany datasets have different training epochs, which are 40, 15, 20, and 20, respectively.}

\subsection{Moving MNIST}\label{sec:mnist}
\textcolor{black}{The Moving MNIST dataset~\cite{srivastava2015unsupervised} contains a total of 15,000 video clips, where each sclip contains 20 frames with 10 frames as history frames and 10 frames as future frames. We divide the training set and test set according to the ratio of 7:3. The size of the frame is $64\times64$, and 2 numbers move randomly in it. These numbers are randomly selected from the static MINIST dataset~\cite{lecun1998gradient}. They start from random positions, random speeds, and random directions, and bounce once they reach the boundary. }

\subsubsection{Visualization of Layer Outputs}  \label{sec:visual}
\textcolor{black}{Fig.~\ref{fig:minist_layer_out_0} shows an example of the inputs, outputs of layers, and outputs of ConvLSTM, SMS-LSTM, TMS-LSTM, and MS-LSTM.} 

\textcolor{black}{We first analyze the horizontal comparison between the two pairs of sub-graphs in Fig.~\ref{fig:minist_layer_out_0}, which is about the difference between the scale-invariant model ConvLSTM (TMS-LSTM) and the scale-variant model SMS-LSTM (MS-LSTM).} \textcolor{black}{(1) Encoder analysis. First, the numbers are getting fatter as can be seen from feature changes (from inputs to layer 2) of all models, indicating that they are all expanding the receptive field to capture motion. Second, SMS-LSTM (MS-LSTM) has a larger receptive field than ConvLSTM (TMS-LSTM) because the numbers in its features are fatter. Third, the numbers 6 and 1 can be distinguished from all features of ConvLSTM (TMS-LSTM) in early moments, and disappear (or became deformed) in later moments (caused by poor performance, the same below). But we cannot distinguish numbers 6 and 1 from layer 2 of SMS-LSTM (MS-LSTM), only abstract outlines. This shows that ConvLSTM (TMS-LSTM) does not learn abstract features, which is contrary to the common knowledge that abstract features generally exist in deep CNN layers. (2) Decoder analysis. From layer 3 to output, SMS-LSTM (MS-LSTM) generates predictions from coarse to fine, where layer 3 gives contours, layer 4 supplements details, and outputs restore resolution. In contrast, ConvLSTM (TMS-LSTM) does not follow this rule, we can always distinguish numbers from their decoders. (3) Model analysis. SMS-LSTM (MS-LSTM) can quickly expand the receptive field to capture motion and can synthesize frames from coarse to fine. ConvLSTM (TMS-LSTM) slowly expands the receptive field, ignores the natural reconstruction process from simple to difficult, and does not learn abstract high-level features from beginning to end. All of these make the performance of ConvLSTM (TMS-LSTM) inferior to SMS-LSTM (MS-LSTM).}

\begin{table}[htbp]
    \centering
	\caption{Ablation study on the Moving MNIST dataset.}
	\resizebox{0.6\linewidth}{!}{
		\centering
		\begin{tabular}{lccccc}
            \toprule
            \centering
            Models & Params\textcolor{black}{$\downarrow$} & Memory\textcolor{black}{$\downarrow$} & FLOPs\textcolor{black}{$\downarrow$} & Time\textcolor{black}{$\downarrow$} & MSE$\downarrow$ \\
            \midrule
            ConvLSTM & \textbf{0.4M} & 3.6G & 34.4G & \textbf{1.9H} & 90.36 \\
            SMS-LSTM & \textbf{0.4M} & \textbf{2.7G} & \textbf{15.1G} & 2.0H & 63.90 \\
            TMS-LSTM & 1.9M & 9.5G & 147.3G & 5.0H & 67.70 \\
            MS-LSTM & 1.9M & 5.2G & 64.5G & 5.0H & \textbf{52.63} \\
            \bottomrule
		\end{tabular}
	} 
	\label{table:mnist-ablation}
\end{table} 

\textcolor{black}{We then analyze the vertical comparison between the two pairs of sub-graphs in Fig.~\ref{fig:minist_layer_out_0}, which is about the difference between the single-cell model ConvLSTM (SMS-LSTM) and the double-cell model TMS-LSTM (MS-LSTM). Overall, the encoding and decoding process of TMS-LSTM (MS-LSTM) is not much different from ConvLSTM (SMS-LSTM). Specifically, the numbers in layer 0 of MS-LSTM are fatter (larger receptive field) than those of SMS-LSTM, which is more obvious in the comparison of layer 1. The reason for this phenomenon is that we introduce MK-LSTM on the basis of SMS-LSTM, which uses a larger convolution kernel with a wider receptive field. However, it is difficult for us to tell which number is fatter between the encoders of TMS-LSTM and ConvLSTM, and it seems that the numbers in layer 0 of TMS-LSTM are even thinner than those of ConvLSTM. This may be because the receptive field of the 3-layer RNN with $3 \times 3$ and $5 \times 5$ convolution is not much different from the receptive field of the 3-layer RNN with $3\times3$ convolution. But if the pooling layer is introduced, these gaps may be further widened. Nevertheless, the long-term predictive ability of TMS-LSTM is still stronger than that of ConvLSTM. ConvLSTM loses the prediction of the contour of the digit 6 at $t=14$ while TMS-LSTM maintains it to the last frame, albeit there is some deformation. Apparently, this is due to the fact that TMS-LSTM uses MK-LSTM with dual temporal memories.}

\begin{table}[htbp]
    \centering
	\caption{Training cost comparison with RNN models on the Moving MNIST dataset.}
	\resizebox{0.6\linewidth}{!}{
		\centering
		\begin{tabular}{lcccc}
			\toprule
			\centering
            Model & Params\textcolor{black}{$\downarrow$} & Memory\textcolor{black}{$\downarrow$} & FLOPs\textcolor{black}{$\downarrow$} & Time\textcolor{black}{$\downarrow$} \\
            \midrule
            \textbf{ConvLSTM}~\cite{shi2015convolutional} & \textbf{0.4M} & \textbf{3.6G} & \textbf{34.4G} & \textbf{1.9H} \\
            PredRNN~\cite{wang2017predrnn} & 0.9M & 6.3G & 69.8G & 4.9H \\
            PredRNN++~\cite{wang2018predrnn++} & 1.4M & 8.8G & 94.7G & 4.5H \\
            PredRNN-V2~\cite{wang2022predrnn} & 0.9M & 7.3G & 70.8G & 7.9H \\
            MIM~\cite{wang2019memory} & 1.8M & 11.0G & 143.0G & 9.5H \\
            MotionRNN~\cite{wu2021motionrnn} & 1.9M & 11.8G & 146.2G & 32.7H \\
            MS-LSTM & 1.9M & 5.2G & 64.5G & 5.0H \\
            \bottomrule
		\end{tabular}
	} 
	\label{table:mnist-cost}
\end{table}

\begin{table}[htbp]
    \centering
	\caption{Quantitative comparison with RNN models on the Moving MNIST dataset.}
	\resizebox{0.45\linewidth}{!}{
		\centering
		\begin{tabular}{lccc}
			\toprule
			\centering
            Model & SSIM$\uparrow$ & MSE$\downarrow$ & MAE$\downarrow$ \\
            \midrule
            ConvLSTM~\cite{shi2015convolutional} & 0.810 & 90.36 & 142.5 \\
            PredRNN~\cite{wang2017predrnn} & 0.845 & 66.17 & 118.0 \\
            PredRNN++~\cite{wang2018predrnn++} & 0.846 & 65.63 & 119.0 \\
            PredRNN-V2~\cite{wang2022predrnn} & 0.855 & 61.57 & 112.9 \\
            MIM~\cite{wang2019memory} & 0.855 & 63.76 & 113.4 \\
            MotionRNN~\cite{wu2021motionrnn} & 0.862 & 59.62 & 108.9 \\
            \textbf{MS-LSTM} & \textbf{0.879} & \textbf{52.63} & \textbf{97.9} \\
            \bottomrule
		\end{tabular}
	} 
	\label{table:mnist-metric}
\end{table}

\begin{figure}[htbp]
	\centering
	\includegraphics[width=0.55\linewidth]{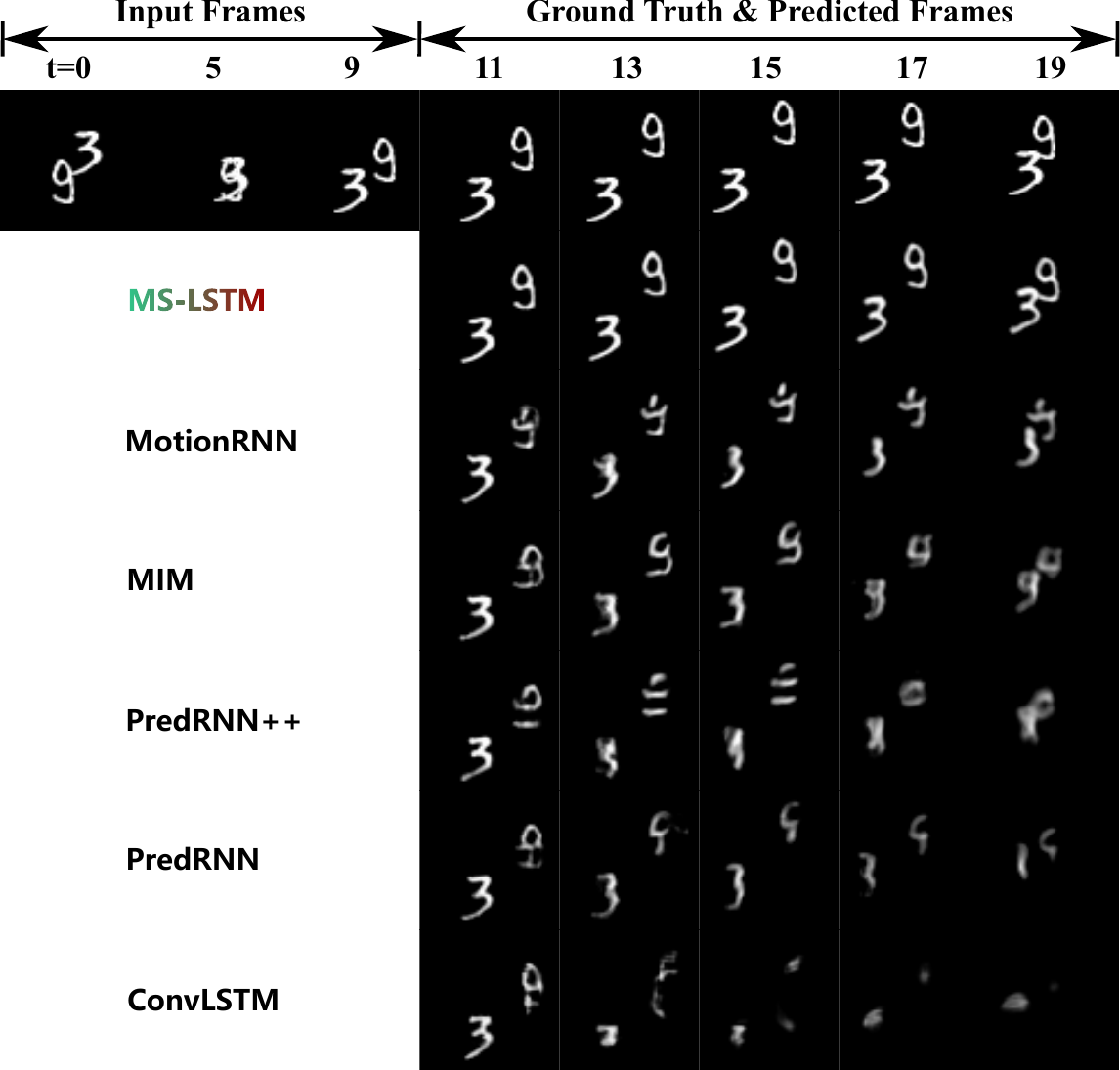}
	\caption{\textcolor{black}{Qualitative comparison on the Moving MNIST dataset.}}
	\label{fig:mnist-demo}
\end{figure}

\subsubsection{Ablation Study} \label{sec:ablation}
As shown in Table \ref{table:mnist-ablation}, both spatial multi-scale LSTM (SMS-LSTM) and temporal multi-scale LSTM (TMS-LSTM) bring performance improvements. Besides, as analyzed in Section~\ref{analysis cost}, the use of pooling does not change the parameters and training time but reduces memory and FLOPs, while the use of large convolution kernels brings additional parameters, memory, FLOPs, and training time.
Since SMS-LSTM brings down the training cost (memory (most tricky) and FLOPs) while TMS-LSTM brings up the training cost (all), we combine SMS-LSTM and TMS-LSTM to construct MS-LSTM to obtain optimal performance without causing excessive training cost.

\subsubsection{\textcolor{black}{Comparison with RNN Models}}
From Table~\ref{table:mnist-cost}, we can see that MS-LSTM has the same amount of parameters as MotionRNN, but the training resources required to train MS-LSTM are much lower than MotionRNN. Specifically, the memory occupation, FLOPs, and training time of MS-LSTM is about \textcolor{black}{1/2, 1/2, and 1/6} of MotionRNN respectively, which is almost less than those of PredRNN. In addition, the quantitative experiments of MS-LSTM also surpassed MotionRNN (Table~\ref{table:mnist-metric}), which proves the significance of the introduction of spatiotemporal multi-scale structure. For example, compared to MotionRNN, the mean square error (MSE) of MS-LSTM is reduced by 11.7\% from 59.62 to 52.63.

\begin{table}[htbp]
    \centering
	\caption{\textcolor{black}{Comparison with competing multiscale non-RNN models on the Moving MNIST dataset.}}
	\resizebox{0.7\linewidth}{!}{
		\centering
		\begin{tabular}{lcccccccc}
			\toprule
			\centering
            Models & Basics & Params\textcolor{black}{$\downarrow$} & Memory\textcolor{black}{$\downarrow$} & FLOPs\textcolor{black}{$\downarrow$} & Time\textcolor{black}{$\downarrow$} & SSIM$\uparrow$ & MSE$\downarrow$ \\
            \midrule
            \textcolor{black}{SimVP}~\cite{gao2022simvp} & CNN & 43.2M & \textbf{4.4G} & 68.5G & 29.4H & 0.734 & 129.2   \\
            \textcolor{black}{Earthformer}~\cite{gao2022earthformer} & Attention & 6.7M & 38.8G & 519.5G & \textbf{3.9H} & 0.787 & 71.0  \\
            \textbf{MS-LSTM} & RNN & \textbf{1.9M} & 5.2G & \textbf{64.5G} & 5.0H & \textbf{0.879} & \textbf{52.6} \\
            \bottomrule
		\end{tabular}
	} 
	\label{table:non-rnn vs}
\end{table}

\textcolor{black}{The prediction task in Fig.~\ref{fig:mnist-demo} becomes extremely difficult due to the heavy overlap of digits 3 and 9 in input frames. The mean and variance of the video undergo drastic changes upon occlusion, indicating the presence of high-order non-stationarity in the sequence~\cite{wang2019memory}.} The image generated by MS-LSTM is more satisfactory than other models, and the ambiguity is not obvious. In fact, we can't even tell the numbers in the last frame generated by other models. We can conclude that MS-LSTM shows great ability in capturing the non-stationary dynamics of complex time-space sequences.

\subsubsection{\textcolor{black}{Comparison with Competing Multiscale Non-RNN Models}}  \label{non-rnn}
\textcolor{black}{At first glance, the CNN-based SimVP~\cite{gao2022simvp} is most similar to our MS-LSTM, which \textcolor{black}{used} the UNet architecture and \textcolor{black}{used} convolution kernels of different sizes. However, SimVP implicitly \textcolor{black}{assumed} that convolutions can simultaneously learn spatiotemporal dynamics, while MS-LSTM explicitly adopts ConvRNN to simultaneously capture spatiotemporal dynamics. In addition, Earthformer~\cite{gao2022earthformer} also \textcolor{black}{used} a UNet-like multi-scale architecture and explicitly \textcolor{black}{used} cuboid attention to capture local and global spatiotemporal dynamics. We compare these two competitors with MS-LSTM in Table~\ref{table:non-rnn vs}. It should be pointed out that we use their official optimal experimental settings. First, MS-LSTM has the fewest parameters and FLOPs, and has the second least memory and training time, both of which are close to the minimum. That is, the training cost of MS-LSTM is almost optimal. Second, MS-LSTM has the largest SSIM (structural similarity,~\cite{wang2004image}) and the smallest MSE, both of which are optimal. Finally, SimVP needs to be trained for 2000 epochs and suffers from severe overfitting. The complex structure of Earthformer requires nearly a thousand lines of code to reproduce, which brings difficulties to its dissemination. In conclusion, MS-LSTM shows superiority in training cost, performance, and training difficulty, and the fundamental reason is its effective multi-scale design and reasonable inductive bias.}

\begin{table}[htbp]
    \centering
	\caption{Training cost comparison on the TaxiBJ dataset.}
	\resizebox{0.55\linewidth}{!}{
		\centering
		\begin{tabular}{lcccc}
			\toprule
			\centering
            Model & Params\textcolor{black}{$\downarrow$} & Memory\textcolor{black}{$\downarrow$} & FLOPs\textcolor{black}{$\downarrow$} & Time\textcolor{black}{$\downarrow$} \\
            \midrule
            \textbf{ConvLSTM}~\cite{shi2015convolutional} & \textbf{0.4M} & \textbf{2.1G} & \textbf{3.2G} & \textbf{0.8H} \\
            TrajGRU~\cite{shi2017deep} & 0.5M & 2.3G & 3.6G & 8.1H \\
            PredRNN~\cite{wang2017predrnn} & 0.9M & 2.4G & 6.4G & 1.9H \\
            PredRNN++~\cite{wang2018predrnn++} & 1.4M & 2.7G & 8.7G & 1.8H \\
            MIM~\cite{wang2019memory} & 1.8M & 2.8G & 13.2G & 3.7H \\
            MotionRNN~\cite{wu2021motionrnn} & 1.9M & 2.9G & 13.5G & 11.8H \\
            CMS-LSTM~\cite{chai2022cms} & 1.1M & 5.6G & 11.5G & 9.4H \\
            \textcolor{black}{MoDeRNN}~\cite{chai2022modernn} & \textcolor{black}{1.5M} & \textcolor{black}{4.9G} & \textcolor{black}{21.4G} & \textcolor{black}{3.8H} \\
            MS-LSTM & 1.9M & 2.3G & 5.9G & 2.0H \\
            \bottomrule
		\end{tabular}
	} 
	\label{table:taxibj-cost}
\end{table}

\begin{table}[htbp]
    \centering
	\caption{Framewise MSE comparison on the TaxiBJ dataset.}
	\resizebox{0.6\linewidth}{!}{
		\centering
		\begin{tabular}{lccccccc}
			\toprule
			\centering
            Model & Frame 1$\downarrow$ & Frame 2$\downarrow$ & Frame 3$\downarrow$ & Frame 4$\downarrow$ \\
            \midrule
            ConvLSTM~\cite{shi2015convolutional} & 0.199 & 0.268 & 0.315 & 0.356 \\
            TrajGRU~\cite{shi2017deep} & 0.193 & 0.264 & 0.305 & 0.352 \\
            PredRNN~\cite{wang2017predrnn} & 0.170 & 0.222 & 0.257 & 0.287 \\
            PredRNN++~\cite{wang2018predrnn++} & 0.162 & 0.200 & 0.234 & 0.268 \\
            MIM~\cite{wang2019memory} & 0.164 & 0.208 & 0.251 & 0.288 \\
            MotionRNN~\cite{wu2021motionrnn} & 0.147 & 0.188 & 0.228 & 0.263 \\
            CMS-LSTM~\cite{chai2022cms} & 0.169 & 0.210 & 0.249 & 0.289 \\
            \textcolor{black}{MoDeRNN}~\cite{chai2022modernn} & \textcolor{black}{0.166} & \textcolor{black}{0.205} & \textcolor{black}{0.240} & \textcolor{black}{0.277} \\
            \textbf{MS-LSTM} & \textbf{0.141} & \textbf{0.173} & \textbf{0.199} & \textbf{0.230} \\
            \bottomrule
		\end{tabular}
	} 
	\label{table:taxibj-metric}
\end{table}

\subsection{TaxiBJ Traffic Flow}\label{sec:taxi}
\textcolor{black}{The TaxiBJ traffic flow dataset, as described in Zhang et al.~\cite{zhang2017deep}, features meteorological and taxicab trajectory data from four time phases in Beijing, gathered by sensors installed in the cars. The trajectory data encompasses two kinds of crowd flows, which are depicted through an $8\times32\times32\times2$ heat map, where the final dimension indicating the intensity of entering and leaving traffic flow in the same region. For our analysis, we employ the entering traffic flow data. Specifically, we use 4 historical frames to forecast 4 future frames, which represents traffic situations for the following two hours. The dataset is split into two parts, following settings outlined in MIM by Wang et al.~\cite{wang2019memory}, where the training set encompasses 19,560 sequences, while the testing set possesses 1,344 sequences.}

\begin{figure}[htbp]
	\centering
	\includegraphics[width=0.5\linewidth]{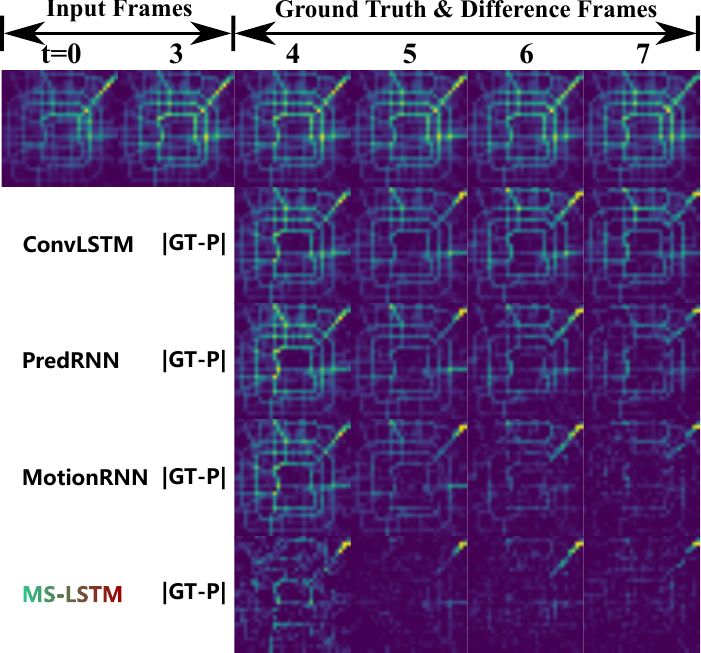}
	\caption{\textcolor{black}{Qualitative comparison on the TaxiBJ dataset. The difference frames between the ground truth and predicted frames are represented by $|$GT-P$|$, where the prediction is optimal when the difference frame is empty.}}
	\label{fig:taxibj-demo}
\end{figure}

\subsubsection{Comparison of the Training Cost and Performance}
\textcolor{black}{Table~\ref{table:taxibj-cost} and Table~\ref{table:taxibj-metric} show the quantitative experimental result.} We can get the same conclusions as above. MS-LSTM requires less training cost but has higher accuracy. The qualitative experimental result is shown in Fig.~\ref{fig:taxibj-demo}, which compares the absolute difference between true and predicted images. Obviously, the prediction of MS-LSTM is the most accurate among all compared models.

\subsubsection{\textcolor{black}{Comparison with Competing Multiscale RNN Models}} \label{sec:taxi-ms-vs}
\textcolor{black}{Lately, two competing multi-scale RNN models have been introduced in 2022, namely CMS-LSTM~\cite{chai2022cms} and \textcolor{black}{MoDeRNN~\cite{chai2022modernn}}. These models have opted for different approaches when it comes to designing multiscale architectures, with CMS-LSTM incorporating multi-patch attention \textcolor{black}{and MoDeRNN utilizing multi-scale convolution kernels.} Unluckily, as can be seen from Table~\ref{table:taxibj-cost} and Table~\ref{table:taxibj-metric}, both of these multiscale designs are quite expensive and inefficient. Even though they have few parameters than MS-LSTM, their FLOPs and memory consumption are nearly more than twice that of MS-LSTM. The attention mechanism with the quadratic complexity used by CMS-LSTM \textcolor{black}{and the oversized convolution kernels (3, 5, and 7) employed by MoDeRNN} contribute to this situation. This also confines them to low-resolution tasks, making it impossible to run them on other high-resolution datasets, such as KTH and Germany. Additionally, both CMS-LSTM and \textcolor{black}{MoDeRNN} achieve poor metrics on the TaxiBJ dataset, failing to surpass the performance of PredRNN++, which indicates that their multiscale designs are ineffective.}


\begin{table}[htbp]
    \centering
	\caption{Training cost comparison on the KTH dataset.}
	\resizebox{0.56\linewidth}{!}{
		\centering
		\begin{tabular}{lcccc}
			\toprule
			\centering
            Model & Params\textcolor{black}{$\downarrow$} & Memory\textcolor{black}{$\downarrow$} & FLOPs\textcolor{black}{$\downarrow$} & Time\textcolor{black}{$\downarrow$} \\
            \midrule
            \textbf{ConvLSTM}~\cite{shi2015convolutional} & \textbf{0.4M} & \textbf{8.3G} & \textbf{137.7G} & \textbf{1.0H} \\
            TrajGRU~\cite{shi2017deep} & 0.5M & 18.1G & 154.2G & 9.4H \\
            PredRNN~\cite{wang2017predrnn} & 0.9M & 18.4G & 279.3G & 2.2H \\
            PredRNN++~\cite{wang2018predrnn++} & 1.4M & 27.7G & 378.7G & 2.8H \\
            MIM~\cite{wang2019memory} & 1.8M & 35.7G & 571.9G & 4.3H \\
            MotionRNN~\cite{wu2021motionrnn} & 1.9M & 38.7G & 584.8G & 12.1H \\
            MS-LSTM & 1.9M & 14.5G & 257.9G & 2.1H \\
            \bottomrule
		\end{tabular}
	} 
	\label{table:kth-cost}
\end{table}

\begin{figure}[htbp]
	\centering
	\subfigure[SSIM]{\includegraphics[width=0.4\linewidth]{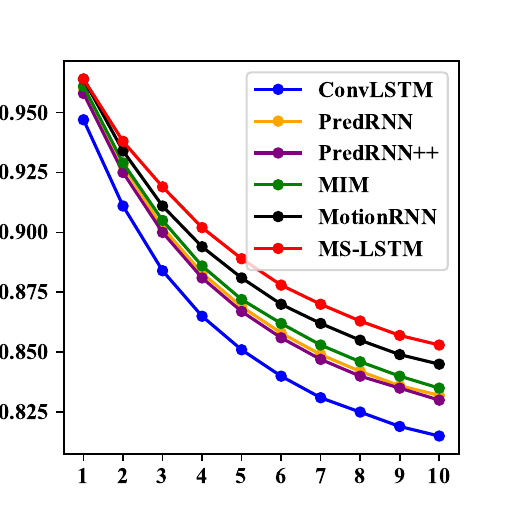}}
        \subfigure[PSNR]{\includegraphics[width=0.4\linewidth]{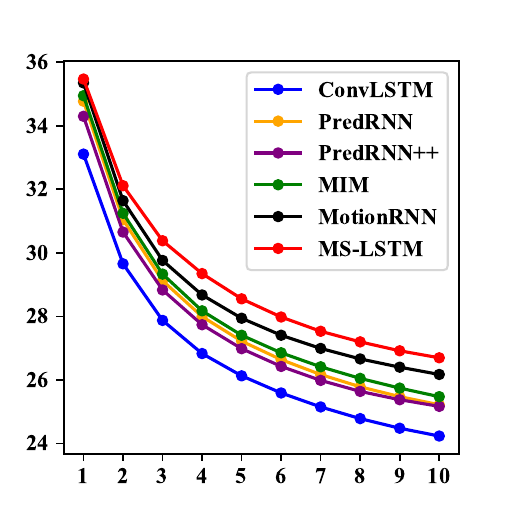}}
	\caption{\textcolor{black}{Framewise SSIM and PSNR comparison on the KTH dataset. The bigger SSIM and PSNR, the better.}}
	\label{fig:kth-metric}
\end{figure}

\begin{figure}[htbp]
	\centering
	\includegraphics[width=0.65\linewidth]{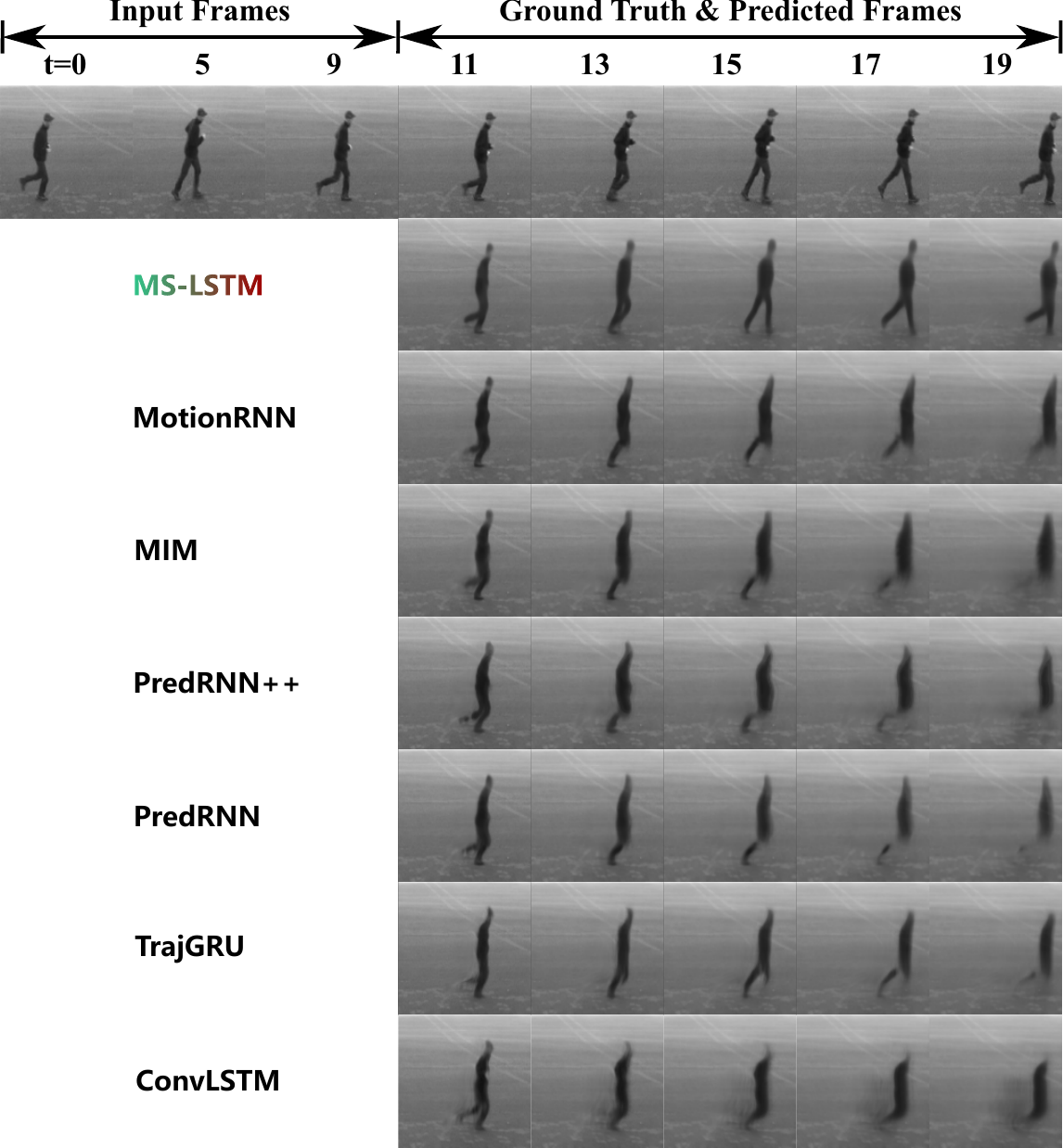}
	\caption{\textcolor{black}{Qualitative comparison on the KTH dataset.}}
	\label{fig:kth-demo}
\end{figure}

\subsection{KTH Human Action}\label{sec:kth}
\textcolor{black}{The KTH dataset~\cite{schuldt2004recognizing} features footage of six human actions (boxing, running, jogging, walking, hand clapping, and hand waving) performed by 25 subjects across four different settings: indoor scenarios, outdoor scenarios, outdoor scenarios with different clothing, and outdoor scenarios with scaling variations. All sequences are captured using a static camera with a 25fps frame rate. The footage is then downsampled to a spatial resolution of $160\times120$ pixels, with an average duration of four seconds. For our study, we focus on the categories of running, jogging, and walking, which primarily involve varying lower body movements. To ensure that human subjects are always visible, we crop frames based on the relevant text in the dataset. We interpolate frames into 128 $\times$ 128 pixels to sustain the running of all models. We use subjects 1-16 for training and subjects 17-25 for testing. During both the training and testing stages, we use a sliding window of 20 for all actions and predict the next 10 frames based on the preceding 10 frames. The stride is set to 10 for jogging and walking, while it is set to 3 for running.}

\subsubsection{Comparison of the Training Cost and Performance}
Table~\ref{table:kth-cost} depicts the training cost comparison on the KTH dataset, demonstrating again that MS-LSTM requires less training cost. \textcolor{black}{We rely on quantitative metrics, including peak signal to noise ratio (PSNR) and SSIM, to gauge the performance of models.} Fig.~\ref{fig:kth-metric} shows the framewise comparisons of SSIM and PSNR, which are obtained by calculating the average of all test sequences at each time step. Apparently, MS-LSTM yields better results than other competing models. Fig.~\ref{fig:kth-demo} also demonstrates this. TrajGRU, PredRNN, PredRNN++, MIM, and MotionRNN only focus on the motion of the left leg, and the right leg is lost in subsequent predictions. Moreover, their predictions are ambiguous, and we can't tell the head from the torso. In contrast, MS-LSTM produces the best predictions, which benefit from the application of multi-scale techniques. \textcolor{black}{The large scales concern the movement of the human body details like legs while the small scales care about the contour of the human body.}

\subsection{Precipitation Nowcasting}\label{sec:germany}
\textcolor{black}{The German radar dataset~\cite{ayzel2020rainnet} contains radar echo maps captured every 5 minutes recorded by 17 Doppler radars between 2006 and 2017. The radar map covers the entire Germany with a resolution of 900$\times$900, which we interpolate to 180$\times$180. Data between 2006 and 2014 are used to optimize the model parameters, while data between 2015 and 2017 are reserved for testing purposes. To make extrapolation more challenging, we sample one radar map every 30 minutes. The model is trained to predict four future frames based on four historical ones, which denote precipitation fluctuations over the next two hours.}

\begin{table}[htbp]
    \centering
	\caption{Training cost comparison on the Germany radar dataset.}
	\resizebox{0.55\linewidth}{!}{
		\centering
		\begin{tabular}{lcccc}
			\toprule
			\centering
            Model & Params\textcolor{black}{$\downarrow$} & Memory\textcolor{black}{$\downarrow$} & FLOPs\textcolor{black}{$\downarrow$} & Time\textcolor{black}{$\downarrow$} \\
            \midrule
            \textbf{ConvLSTM}~\cite{shi2015convolutional} & \textbf{0.4M} & \textbf{6.8G} & \textbf{100.3G} & \textbf{0.8H} \\
            TrajGRU~\cite{shi2017deep} & 0.5M & 14.0G & 112.4G & 7.7H \\
            PredRNN~\cite{wang2017predrnn} & 0.9M & 14.3G & 203.5G & 1.9H \\
            PredRNN++~\cite{wang2018predrnn++} & 1.4M & 21.1G & 275.9G & 2.3H \\
            MIM~\cite{wang2019memory} & 1.8M & 27.1G & 416.7G & 3.8H \\
            MotionRNN~\cite{wu2021motionrnn} & 1.9M & 29.4G & 426.1G & 10.4H \\
            PrecipLSTM~\cite{ma2022preciplstm} & 1.8M & 38.3G & 398.5G & 14.8H \\
            MS-LSTM & 1.9M & 14.8G & 260.6G & 1.8H \\
            \bottomrule
		\end{tabular}
	} 
	\label{table:germany-cost}
\end{table}

\begin{table}[htbp]
    \centering
	\caption{Quantitative comparison on the Germany radar dataset.}
	\resizebox{0.7\linewidth}{!}{
		\centering
		\begin{tabular}{lcccccc}
			\toprule
			\centering
            Model & CSI-0.5$\uparrow$ & CSI-2$\uparrow$ & CSI-5$\uparrow$ & HSS-0.5$\uparrow$ & HSS-2$\uparrow$ & HSS-5$\uparrow$ \\
            \midrule
            ConvLSTM~\cite{shi2015convolutional} & 0.341 & 0.199 & 0.079 & 0.490 & 0.315 & 0.138 \\
            TrajGRU~\cite{shi2017deep} & 0.378 & 0.210 & 0.082 & 0.531 & 0.331 & 0.144 \\
            PredRNN~\cite{wang2017predrnn} & 0.380 & 0.231 & 0.092 & 0.533 & 0.358 & 0.158 \\
            PredRNN++~\cite{wang2018predrnn++} & 0.382 & 0.234 & 0.091 & 0.536 & 0.364 & 0.158 \\
            MIM~\cite{wang2019memory} & 0.380 & 0.225 & 0.087 & 0.534 & 0.352 & 0.153 \\
            MotionRNN~\cite{wu2021motionrnn} & 0.387 & 0.229 & 0.085 & 0.541 & 0.357 & 0.149 \\
            PrecipLSTM~\cite{ma2022preciplstm} & 0.392 & 0.234 & 0.088 & 0.556 & 0.364 & 0.152 \\
            \textbf{MS-LSTM} & \textbf{0.405} & \textbf{0.243} & \textbf{0.096} & \textbf{0.562} & \textbf{0.378} & \textbf{0.164} \\
            \bottomrule
		\end{tabular}
	} 
	\label{table:germany-metric}
\end{table}

\begin{figure*}[htbp]
	\centering
	\subfigure[CSI]{\includegraphics[width=0.49\linewidth]{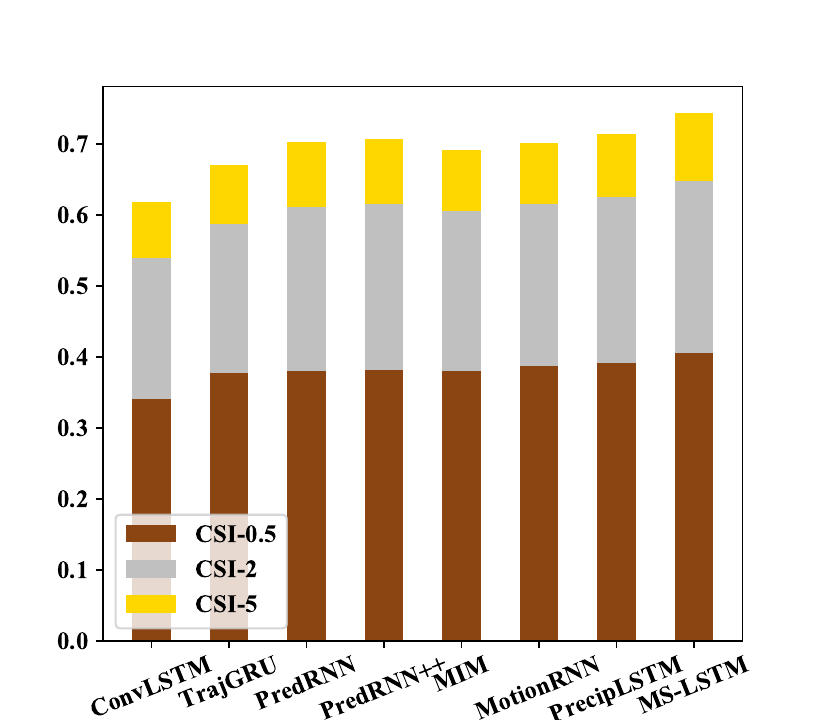}}
        \subfigure[HSS]{\includegraphics[width=0.49\linewidth]{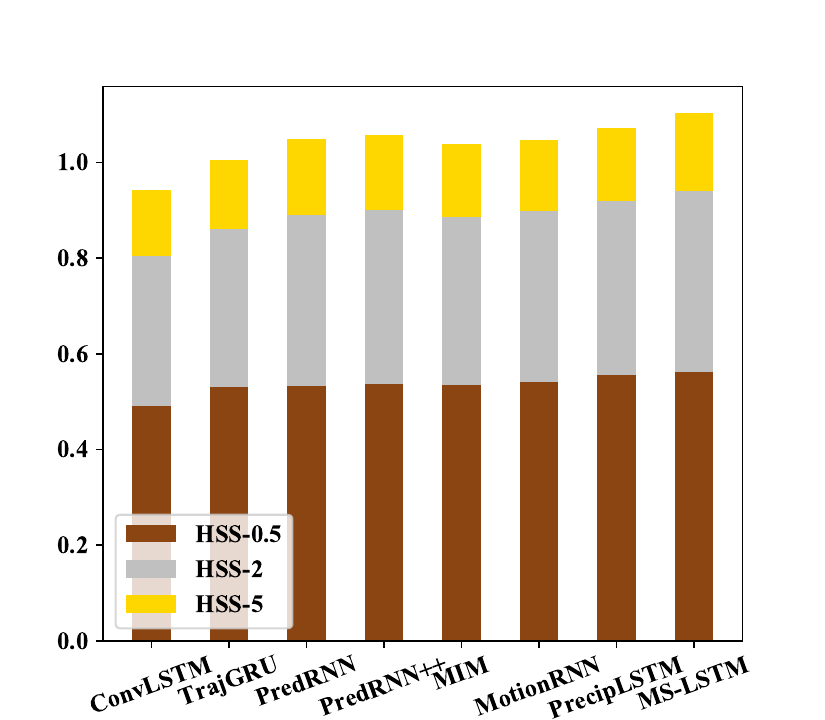}}
	\caption{\textcolor{black}{Accumulate CSI and HSS comparison on the Germany radar dataset.}}
	\label{fig:germany-accumulate}
\end{figure*}

\begin{figure}[htbp]
	\centering
	\includegraphics[width=0.55\linewidth]{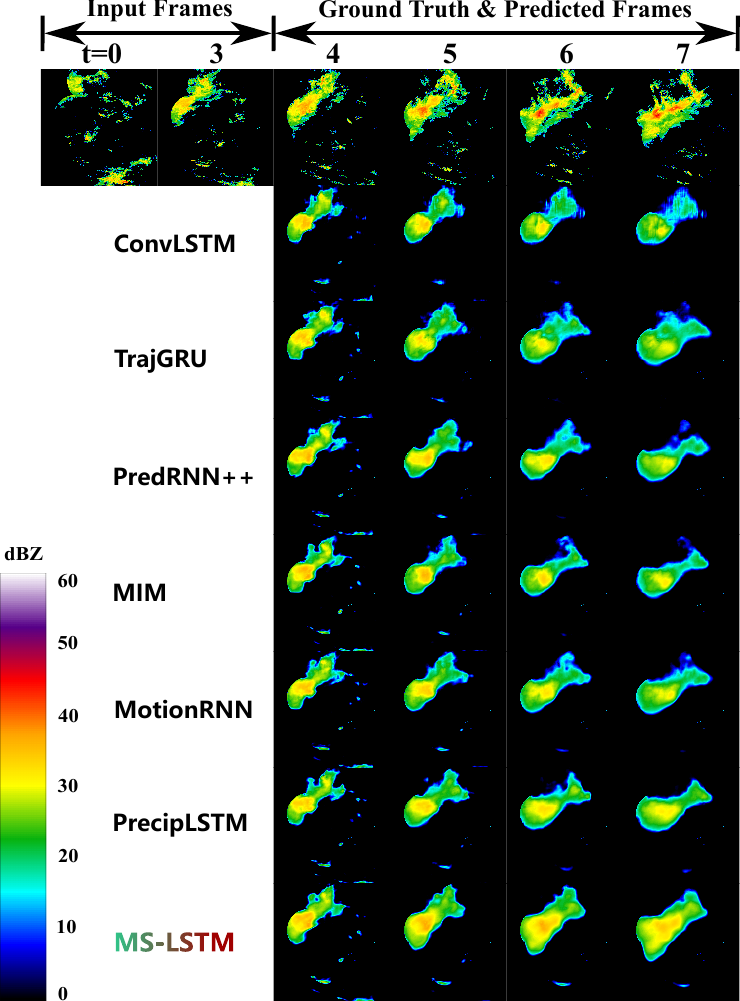}
	\caption{\textcolor{black}{Qualitative comparison on the Germany radar dataset.}}
	\label{fig:germany-demo}
\end{figure}

\subsubsection{Comparison of the Training Cost and Performance}

We report comparisons of training cost and precipitation metrics between our model and competing models in Table~\ref{table:germany-cost}, Table~\ref{table:germany-metric}, and Fig.~\ref{fig:germany-accumulate}, and the results all demonstrate one thing: MS-LSTM is optimal. It is worth noting that we only report the critical success index (CSI)~\cite{shi2017deep} and Heidke skill score (HSS) metrics~\cite{shi2017deep} of models at the 0.5, 2, and 5 mm/h thresholds. This is mainly because Germany has a temperate maritime climate, where the precipitation is mostly light rain and light showers, with almost no heavy rain. Fig.~\ref{fig:germany-demo} depicts one precipitation movement process that diffuses from northwestern to central Germany. As can be seen from the figure, each model predicts differently for the last frame (precipitation after 2 hours). Among them, the prediction of MS-LSTM is closest to the true value, which is thanks to the usage of the large convolution kernel to strengthen the multi-scale architecture to capture fast motion. In addition, although precipitation exhibits long-tail distribution and heavy rain is rare, MS-LSTM still focuses on heavier rain.

\section{Conclusion and Future Work}\label{conclusion and future work}
\textcolor{black}{In order to refresh the performance on the video prediction task, the previous RNN model becomes wider and deeper, which brings unbearable training cost. Different from these practices, this paper proposes a new model MS-LSTM from the perspective of scale. MS-LSTM is based on three orthogonal multi-scale designs, namely depth, downsampling, and multi-kernel. We theoretically analyze the training cost and performance of MS-LSTM and its components to enrich the interpretability of MS-LSTM, which is demonstrated in experiments on Moving MNIST. In order to verify the superiority of MS-LSTM over other models, we conducted extensive comparative experiments on four video datasets. The comparison models include advanced common ConvRNN models, competitive multiscale ConvRNN models, and competitive multiscale non-RNN models. The results show that MS-LSTM has higher performance but less training cost.}

\textcolor{black}{In addition to using pooling to build a efficient multi-scale structure, another technique to reduce training costs (mainly video memory) is to use patch, which divides the image into small blocks of the same size. These patches are stitched together in the channel dimension as a tensor input into the neural network, and restored to the original image resolution when output. ConvLSTM first \textcolor{black}{used} this approach, mainly due to the backwardness of the training equipment at that time. Although the patch brings a large drop in training cost, it also leads to a significant drop in model performance and may cause raster effects in the predicted image. Therefore, this design is not considered in this paper. However, the use of patch is the fundamental reason why the Transformer architecture can be transferred from natural language processing to vision. ViT \textcolor{black}{used} patch to greatly reduce the quadratic complexity of attention. MaskViT \textcolor{black}{attempted} to perform cloze tasks on patches to force the model to predict new frames. In short, patch can reduce training usage and may become a new paradigm for video prediction, but further improvement and research are needed in the future.}

\section{Acknowledgments}
This work is partly supported by the National Key R\&D Program of China under Grant No. 2021ZD0110900, the National Natural Science Foundation of China under Grant No. 62106061, 61972114, the Fundamental Research Funds for the Central Universities under Grant No. FRFCU5710010521, the Research and Application of Key Technologies for Intelligent Farming Decision Platform, An Open Competition Project of Heilongjiang Province (China) under Grant No. 2021ZXJ05A03, the Key Research and Development Program of Heilongjiang Province under Grant No. 2022ZX01A22, the National Natural Science Foundation of Heilongjiang Province under Grant No. YQ2019F007.

\bibliography{ms-lstm}
\bibliographystyle{elsarticle-num}   

\end{sloppypar}
\end{document}